\pdfoutput=1
\documentclass[journal]{IEEEtran}

\usepackage{amsmath,amssymb,amsfonts}
\usepackage{amsthm}
\usepackage{cite}
\usepackage{xurl}
\usepackage[dvipsnames]{xcolor}
\usepackage[hidelinks]{hyperref}
\usepackage{graphicx}
\usepackage[caption=false]{subfig}
\usepackage{booktabs}
\usepackage{array}
\usepackage{multirow}
\usepackage{stfloats}       
\usepackage{orcidlink}
\usepackage{tikz}
\usetikzlibrary{arrows.meta,positioning,calc}

\makeatletter
\def\section{\@startsection{section}{1}{\z@}{2.2ex plus 0.3ex minus 0.6ex}{0.7ex plus 0.2ex minus 0ex}{\normalfont\normalsize\centering\scshape}}
\def\subsection{\@startsection{subsection}{2}{\z@}{2.2ex plus 0.3ex minus 0.6ex}{0.6ex plus 0.2ex minus 0ex}{\normalfont\normalsize\itshape}}
\makeatother

\theoremstyle{definition}

\theoremstyle{plain}
\newtheorem{theorem}{Theorem}
\newtheorem{lemma}{Lemma}
\newtheorem{proposition}{Proposition}
\newtheorem{corollary}{Corollary}
\newtheorem{assumption}{Assumption}
\theoremstyle{remark}

\newcommand{\Tset}{\mathcal{T}}
\newcommand{\Sset}{\mathcal{S}}
\newcommand{\Aset}{\mathcal{A}}
\newcommand{\Nset}{\mathcal{N}}

\newcommand{\R}{\mathbb{R}}
\newcommand{\Gone}{\mathbb{G}_1}
\newcommand{\Gtwo}{\mathbb{G}_2}

\newcommand{\negl}{\mathsf{negl}}

\usepackage{needspace}
\usepackage{threeparttable}
\usepackage[final]{microtype}
\begin{document}

\bstctlcite{BSTcontrol}

\title{Fair and Efficient Distributed EV Charging Coordination Using
Threshold Cryptography and Machine Learning}

\author{Md. Mainul Islam\orcidlink{0000-0002-9686-5994},~\IEEEmembership{Graduate Student Member,~IEEE,}~
        Muhammad Ismail\orcidlink{0000-0002-8051-9747},~\IEEEmembership{Senior Member,~IEEE,}~
        Hasan Kurban\orcidlink{0000-0003-3142-2866},~Abdulrahman Takiddin\orcidlink{0000-0003-4793-003X},~\IEEEmembership{Member,~IEEE,}~
        and Erchin Serpedin\orcidlink{0000-0001-9069-770X},~\IEEEmembership{Fellow,~IEEE}%
        \IEEEcompsocitemizethanks{
            \IEEEcompsocthanksitem Md. Mainul Islam and Erchin Serpedin  are with the Department of Electrical and Computer Engineering, Texas A\&M University, College Station, TX 77843, USA (e-mail: mdmainul11@tamu.edu, eserpedin@tamu.edu). Muhammad Ismail is with the  Cybersecurity Education, Research, and Outreach Center (CEROC) and Department of Computer Science,  Tennessee Tech University, Cookeville, TN 38505, USA (e-mail:  mismail@tntech.edu). Hasan Kurban is with the College of Science and Engineering, Hamad Bin Khalifa University, Doha, Qatar (e-mail: hkurban@hbku.edu.qa). Abdulrahman Takiddin is with the Department of Electrical and Computer Engineering, FAMU-FSU College of Engineering, Florida State University, Tallahassee, FL 32310 USA (e-mail: a.takiddin@fsu.edu). Corresponding author: Md Mainul Islam.
        }
     \thanks{This work has been submitted to the IEEE for possible publication. Copyright may be transferred without notice, after which this version may no longer be accessible.}
}

\maketitle

\begin{abstract}
Distributed coordination of electric vehicle (EV) charging with the alternating direction method of multipliers (ADMM) keeps EV constraints private but assumes that every participant follows the algorithm. This assumption fails in practice, where a biased operator or a man-in-the-middle (MITM) attacker can send inconsistent signals to different EVs, and compromised EVs can deviate from their prescribed updates. We first prove that attacks by such arbitrarily malicious, i.e., Byzantine, participants do not prevent convergence and therefore remain undetected, and derive the most damaging attack in closed form. We then propose a framework that combines a threshold-signed broadcast with upload recomputation, which detects every deviation without false alarms at linear instead of quadratic communication cost. To keep the optimization running, each affected EV recovers locally with a certified, fleet-aware machine learning price forecast, and we prove that the optimization remains convergent with a recovery cost bounded by the shape of the forecast error. Simulation results with real-world load, price, and charging data show that a biased operator raises the annual cost of disadvantaged EVs by 132.7\% and lowers that of advantaged EVs by 49.3\% without changing the aggregate load, whereas the defense limits all cost changes to 2.6\%. An external MITM attacker raises the cost of its victims by 57.6\%, the peak load by 5.8\%, and the grid cost by 15.8\%, whereas the defense keeps them within 1.1\%, 0.2\%, and 0.3\% of benign operation. With up to 70\% of the EVs attacked, the recovery converges in at least 97.8\% of the runs.
\end{abstract}


\section{Introduction}
Electric vehicles (EVs) are central to the decarbonization of transportation, and coordinated charging provides demand flexibility that facilitates the integration of renewable generation~\cite{Needell2023BeneficialCharging}. As adoption grows, charging demand becomes concentrated in specific locations and time windows~\cite{Zhan2025_LargeScaleEVUsage}, and uncoordinated charging can coincide with existing peaks, increasing net demand, feeder congestion, and transformer loading~\cite{Powell2022NatureEnergy}.

These effects motivate optimization-based charging, which can be centralized or distributed. Centralized methods compute globally optimal schedules~\cite{HybridFleetLoadScheduling,TwoStageAggregatorReliability,Kececi2021CheatingCustomers,Fahim2025DynamicPlanningGCNN,Alizadeh2025TII,Alizadeh2026TTE,Alizadeh2025ETFG}, but they require the operator to collect all EV constraints, some of which, such as the state of charge and usage patterns, are sensitive, and to solve large problems repeatedly. Distributed methods instead keep private inputs local and exchange only low-dimensional coordination signals~\cite{ADMMFastChargeDegradation,UncertainDepartureLoadFlattening,Nimalsiri2020SurveyDistributedCharging}. The alternating direction method of multipliers (ADMM) is popular because it scales to large fleets and converges for convex problems~\cite{Boyd2011ADMM}.

\begin{table*}[!t]
\centering
\begin{threeparttable}
\caption{Comparison With Prior Work on Attacks in Distributed Optimization}
\label{tab:related}
\renewcommand{\arraystretch}{1.15}
\scriptsize
\setlength{\tabcolsep}{2pt}
\begin{tabular*}{\textwidth}{@{\extracolsep{\fill}}llcccclc@{}}
\toprule
\textbf{Work} & \textbf{Byzantine parties} & \textbf{Stealth}\tnote{a} & \textbf{Detection} & \textbf{Trusted channel}\tnote{b} & \textbf{Liveness}\tnote{c} & \textbf{Guarantees} & \textbf{Messages}\tnote{d}\\
\midrule
Attacks by participating EVs~\cite{Fard2023ISIE,FakouriFard2025Targeted} & EVs & Yes & None & N/A & No & Attack analysis & N/A\\
Detection of deviating aggregators~\cite{PerezDiaz2020Catching} & Aggregators & Partly & Statistical & N/A & No & Empirical & N/A\\
Byzantine-robust gradient aggregation~\cite{Blanchard2017Krum,Yin2018TrimmedMean} & Workers & No & Filtering & No & Partly & Statistical error bound & $O(N)$\\
Resilient distributed optimization~\cite{Sundaram2019Adversarial,Yemini2025Resilient} & Neighbors & No & Filtering & Yes (P2P) & Partly & Approximate convergence & $O(L)$\\
Trust-based node isolation~\cite{Qu2026Resilient} & EVs, links & Partly & Threshold-based & Yes (P2P) & Partly & Convergence after isolation & $O(L)$\\
Blockchain-based coordination~\cite{Baza2019BlockchainCC} & Coordinator, participants & No & Consensus & Yes (ledger) & No & Consensus & $O(N^2)$\\
\textbf{Proposed framework} & Coordinator, MITM, EVs & Yes & Deterministic & No & Yes & Consistency, convergence & $O(N)$\\
\bottomrule
\end{tabular*}
\begin{tablenotes}[para]\scriptsize
\item[a] Attacks that preserve convergence are analyzed.
\item[b] P2P links or a ledger in addition to the coordinator.
\item[c] The optimization provably continues and converges under attack.
\item[d] Per iteration; $L$: number of P2P links. N/A: not applicable.
\end{tablenotes}
\end{threeparttable}
\end{table*}

The guarantees of ADMM, however, rest on the implicit assumption that all participants follow the algorithm. The operator must send the same coordination signal to every EV, and every EV must compute its prescribed update. This assumption is difficult to ensure in deployed charging systems. The coordination signal passes through chargers, site controllers, and cloud gateways, each of which terminates and re-originates the secured session~\cite{OCPP201,ISO15118_20}, and these devices, as well as the EV-side software that computes the local updates, have documented vulnerabilities~\cite{Acharya2020Cybersecurity,Johnson2022Review}. Consequently, a biased operator, or a man-in-the-middle (MITM) attacker that controls an intermediary, can send mutually inconsistent coordination signals to different EVs, which we call an \emph{inconsistency attack} and which is known in distributed computing as \emph{equivocation}; likewise, a compromised EV can deviate from its prescribed update. Participants that may deviate arbitrarily from a protocol, rather than merely failing to respond, are called \emph{Byzantine}~\cite{Lamport1982Byzantine}. As shown in Section~\ref{sec:threat_model}, an iteration under an inconsistency attack still converges, and the attack distorts the victims' schedules and payments without any observable anomaly.

Detecting inconsistency attacks is difficult because EVs communicate only with the operator. The classical remedy lets the receivers compare or monitor the messages that they receive~\cite{Bracha1987Reliable,Qu2026Resilient}; for EV charging, this requires peer-to-peer (P2P) links among the EVs and $O(N^2)$ messages per iteration for $N$ EVs, which limits scalability to large fleets. A trusted online channel, such as a public bulletin board, would instead defeat the purpose of coordinating through an operator. A practical defense must therefore satisfy three requirements: \emph{consistency}, i.e., every EV receives the same signal and follows its prescribed update even if the operator, a MITM attacker, or an EV is Byzantine; \emph{scalable detection}, i.e., every violation is detected with certainty, without false alarms and without P2P links, ledgers, or a bulletin board; and \emph{liveness}, i.e., the iterative optimization continues to progress and converges, with provable cost guarantees, even when some EVs detect an attack.

Neither cryptography nor machine learning (ML) can meet these requirements alone. Transport Layer Security (TLS), which protects each communication hop, cannot detect tampering at an intermediary that terminates the session, an operator can sign whatever it sends, and detection alone leaves the affected EV without a valid signal, and the optimization stalls. ML can estimate the correct signal but cannot detect manipulation with certainty, and each false alarm would replace a correct signal with an estimate. Blockchain-based designs avoid trusting the operator, but their consensus protocols exchange $O(N^2)$ messages per decision and replicate data at every node~\cite{Baza2019BlockchainCC}. This paper therefore integrates cryptographic detection with ML-based self-recovery. A threshold-signed broadcast detects inconsistent signals without any inter-EV communication, and every EV that detects an attack continues charging with a schedule that it computes locally from a certified ML price forecast. No message is exchanged among EVs, communication remains $O(N)$ per iteration, as in plain ADMM, and the optimization remains live under attack.

\subsection{Related Work and Research Gaps}
Prior studies on charging coordination reduce cost and flatten demand under network limits, battery degradation, and uncertain departures~\cite{ADMMFastChargeDegradation,TwoStageAggregatorReliability,HybridFleetLoadScheduling,UncertainDepartureLoadFlattening}. Planning studies site and size storage and charging stations with genetic algorithms and deep reinforcement learning~\cite{Alizadeh2025TII}, extend this approach to coupled power and transportation networks~\cite{Alizadeh2026TTE}, and incorporate vehicle-to-grid (V2G) capability~\cite{Alizadeh2025ETFG}. These studies assume a single planner that holds all data and trustworthy signals.

Security studies have addressed dishonest customers, privacy-preserving authentication, dynamic-charging protection, and blockchain-assisted charging and energy trading~\cite{Kececi2021CheatingCustomers,Abouyoussef2025SecureSemiDecentralizedDWC}, as well as threats to the charging infrastructure~\cite{Acharya2020Cybersecurity,Johnson2022Review}. Attacks on charging coordination have been attributed to participants, namely EVs or aggregators that misreport their data or deviate from their updates~\cite{Fard2023ISIE,FakouriFard2025Targeted,PerezDiaz2020Catching,Shafee2020LyingEV}, or to outsiders that inject false measurements~\cite{Fahim2024GAETTE}. In decentralized ADMM, Perez-Diaz~\emph{et al.}~\cite{PerezDiaz2020Catching} showed that self-interested aggregators can steer the iteration to a non-optimal solution that benefits them and detected such manipulation statistically. In a fully distributed, consensus-based scheduler, Qu~\emph{et al.}~\cite{Qu2026Resilient} detected false data injection, denial-of-service, and replay attacks through mutual observations among neighboring EVs, isolated low-trust nodes, and reset prices so that the remaining EVs converge to the optimum of the reduced problem; this defense requires P2P communication and observation links among the EVs, and its threshold-based detection can isolate honest nodes under severe disturbances. These defenses are learned, statistical, or threshold-based detectors whose probabilistic decisions degrade under poisoning and evasion attacks~\cite{Takiddin2021Poisoning}.

Beyond EV charging, robust aggregation rules, such as Krum~\cite{Blanchard2017Krum} and the trimmed mean~\cite{Yin2018TrimmedMean}, protect a trusted server against faulty workers, and resilient consensus and optimization methods filter or trust-weight the values received from neighbors in P2P networks~\cite{Sundaram2019Adversarial,Duan2019Resilient,Yemini2025Resilient}. Blockchain-based designs can replace the coordinator with a smart contract~\cite{Baza2019BlockchainCC}. Inconsistent broadcasting by a sender is classically prevented either by receivers that echo the received messages to one another~\cite{Bracha1987Reliable}, which requires P2P links and $O(N^2)$ messages, or by a group of designated signers that certifies each message with a threshold signature, i.e., a signature that only a sufficiently large subset of the signers can produce jointly~\cite{Cachin2001Broadcast}. The signing key of such a scheme is normally generated over a broadcast channel among the signers~\cite{Gennaro2007DKG}.

Table~\ref{tab:related} compares prior work on attacks in distributed optimization with respect to the three requirements and, together with the studies above, reveals three limitations. First, prior studies of attacks on charging coordination assume a trustworthy coordinator, and the adversary is either a participant~\cite{Fard2023ISIE,FakouriFard2025Targeted,PerezDiaz2020Catching,Shafee2020LyingEV} or an outsider~\cite{Fahim2024GAETTE}. Inconsistency attacks by the coordinator, or by a MITM attacker on its downlink, have not been modeled, and the damage of attacks that preserve convergence and thus evade detection has not been characterized. Second, as Table~\ref{tab:related} shows, existing defenses of distributed optimization are probabilistic, trust the server, or require P2P links or a ledger. Statistical and threshold-based detectors can raise false alarms, which would exclude honest participants, whereas filtering discards outlying values without identifying their source and therefore guarantees only approximate convergence; in particular, detecting inconsistent signals without trusting the coordinator has until now required inter-EV communication with $O(N^2)$ messages~\cite{Bracha1987Reliable}. Third, existing defenses end at detection, exclusion, or agreement, or, as in~\cite{Qu2026Resilient}, restore convergence only for the EVs that remain after the suspected nodes are removed. None of them keeps the attacked EVs themselves in a coordinator-based optimization with a convergence guarantee, and none bounds their cost relative to alternative recovery rules, as this paper does.

\subsection{Contributions}
The main contributions of this paper are as follows:
\begin{itemize}\setlength{\itemsep}{1pt}\setlength{\topsep}{2pt}
    \item \emph{Two-sided threat model:} Unlike prior studies, which trust the coordinator, we consider a biased coordinator and a MITM attacker that send inconsistent signals in the iterative loop, together with EVs that deviate from their updates. We prove that such attacks converge to a perturbed optimum without any observable anomaly and derive their damage and the most damaging attack in closed form (Proposition~\ref{prop:stealth}, Corollary~\ref{cor:damage}).
    \item \emph{Scalable deterministic detection:} Our certified broadcasts, whose key is generated through the untrusted operator without a trusted dealer, and the recomputation of uploads detect every attack without false alarms and at $O(N)$ communication cost, compared with $O(N^2)$ for P2P echo broadcast (Lemma~\ref{lem:dkg}, Corollary~\ref{cor:detection}).
    \item \emph{Reduction to erasures:} We reduce every attack to an erasure, i.e., EVs that stop following the signal, which links cryptography to the optimization analysis (Corollary~\ref{cor:content}).
    \item \emph{Liveness and convergence under attack:} We prove that local recovery from a certified, fleet-aware ML forecast converges and remains feasible from any attack onset, with a cost bounded by the shape of the forecast error and near-optimal among forecast-based rules (Theorems~\ref{thm:conv_recovery}--\ref{thm:baselines}).
    \item \emph{Experimental validation:} We simulate the optimization model with real-world load, price, and EV charging data under uncoordinated, benign, attacked, and post-attack recovery conditions, and coordination alone lowers the EV energy cost on every day, by 10.4\% in the median. Undefended attacks raise the annual costs of targeted EVs by up to 132.7\% and the peak load by 5.8\%, whereas the defense keeps all costs and the peak within 2.6\% of benign operation. Its recovery converges in at least 97.8\% of the runs with up to 70\% of the EVs attacked, far more often than alternative rules.
\end{itemize}

Sections~\ref{sec:problem_formulation}--\ref{sec:proposed_secure_framework} present the system model, the threat model, and the proposed framework, Section~\ref{sec:implementation_results} evaluates the framework, Section~\ref{sec:limitations} discusses its limitations, and Section~\ref{sec:conclusion} concludes the paper. All proofs are given in the Appendix.

\section{System Model and Distributed Coordination}
\label{sec:problem_formulation}

\subsection{System Model}
\label{subsec:system_model}

\begin{figure}[t]
    \centering
    \includegraphics[width=0.99\linewidth]{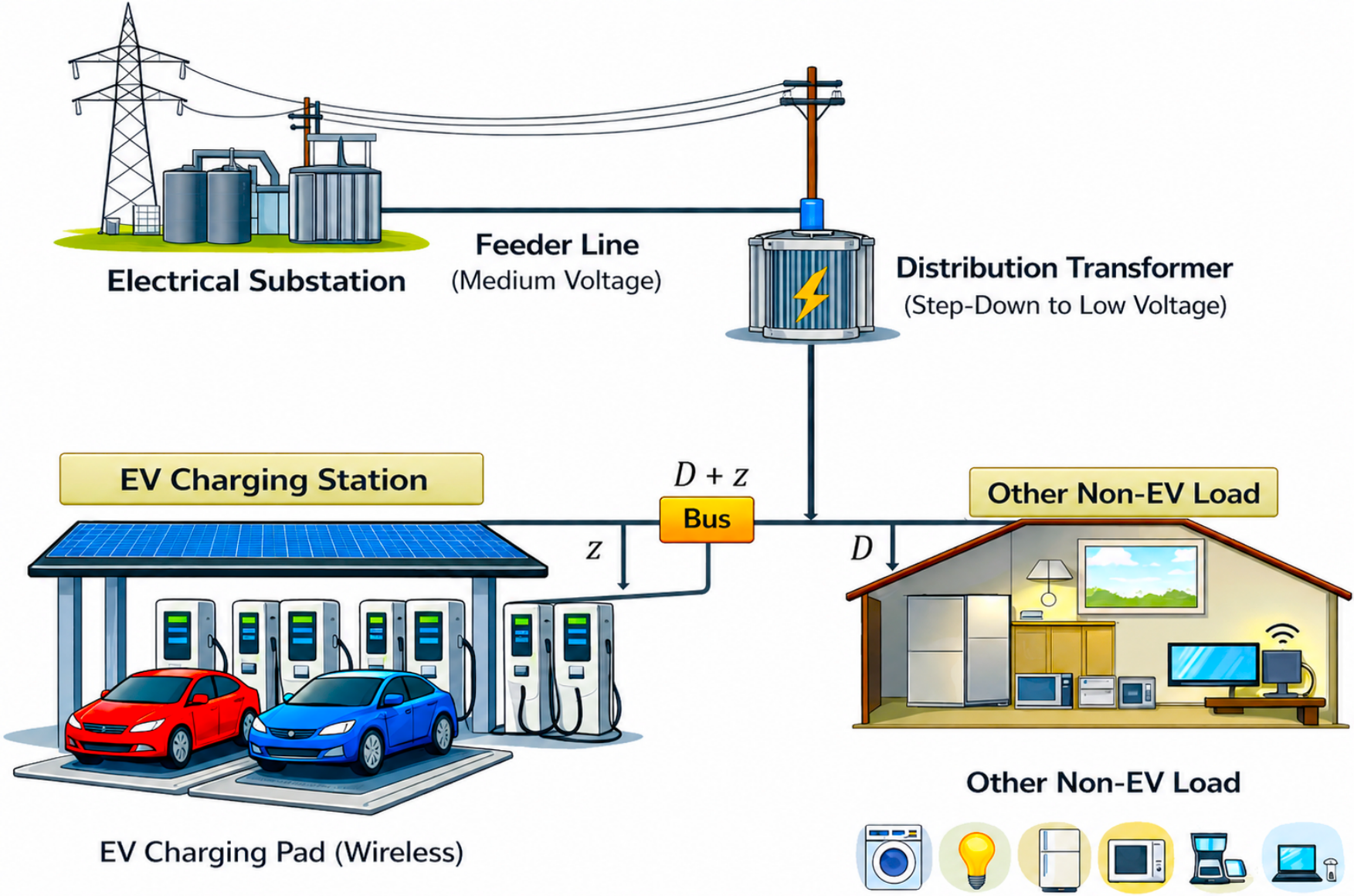}
    \caption{Charging station and non-EV load served by a common bus.}
    \label{fig:CS-Model}
\end{figure}

As shown in Fig.~\ref{fig:CS-Model}, a distribution transformer supplies a common low-voltage bus that serves both the EV charging station (CS) and the non-EV load. Let $D(t)$ denote the non-EV demand in time slot $t$ and $z(t)$ the aggregate EV charging demand, so that the total bus load is $D(t)+z(t)$.

Consider $N$ EVs, indexed by $\Nset=\{1,\ldots,N\}$, that a CS operator coordinates day-ahead over the horizon $\Tset=\{1,\ldots,T\}$, where $T$ is the number of time slots and $\Delta t$ is the slot duration. EV $i$ is available for charging during the slots $\Tset_i\subseteq\Tset$ between its declared arrival and departure, requires the energy $E_i$, and has the per-slot charging limit $p_i^{\max}$; the CS has the power rating $P_{\mathrm{CS}}^{\max}$. Let $x_i(t)$ be the charging power of EV $i$ in slot $t$, so that $z(t)=\sum_{i=1}^{N}x_i(t)$. The operator faces the grid tariff $c(t)$ and a network-stress weight $q(t)>0$ that penalizes a high total load $D(t)+z(t)$. The grid cost is $J(\mathbf{z})\triangleq\sum_{t\in\Tset}\big[q(t)\big(D(t)+z(t)\big)^2+c(t)z(t)\big]\Delta t$, and each EV has a local preference cost $\varphi_i(\mathbf{x}_i)$ defined below. The problem minimizes the total cost under EV and station constraints:
\begin{align}
\min_{\{\mathbf{x}_i\},\mathbf{z}}\quad
&{J(\mathbf{z})+\sum_{i=1}^{N}\varphi_i(\mathbf{x}_i)}
\label{eq:obj_site}\\
\text{s.t.}\quad
&z(t)=\sum_{i=1}^{N}x_i(t),\quad
0\le z(t)\le P_{\mathrm{CS}}^{\max},
\label{eq:coupling_sum_site}\\
&0\le x_i(t)\le p_i^{\max}\ {(t\in\Tset_i)},\quad {x_i(t)=0\ (t\notin\Tset_i)},
\label{eq:rate_window}\\
&\sum_{t\in\Tset}x_i(t)\Delta t=E_i,
\label{eq:site_rating}
\end{align}
for all $i$ and $t$. The constraints enforce aggregate-load consistency, the CS capacity, the charging-rate limits and availability of the EVs, and their energy requirements.

The preference cost of EV $i$ is
\begin{equation}
\varphi_i(\mathbf{x}_i)=\omega\sum_{t\in\Tset_i}\big(t-t_i^{\mathrm{a}}\big)x_i(t)\Delta t+\frac{\beta}{2}\big\|\mathbf{x}_i-\mathbf{x}_i^{\mathrm{p}}\big\|_2^2,
\label{eq:preference}
\end{equation}
where $t_i^{\mathrm{a}}$ is its arrival slot, $\omega\ge0$ is the cost of delaying one unit of energy by one slot, and $\beta\ge0$ weights the deviation from the preferred schedule $\mathbf{x}_i^{\mathrm{p}}$, in which EV $i$ charges at $p_i^{\max}$ from arrival until $E_i$ is delivered. The first term favors early charging at the rated power, which yields realistic charging durations of about $E_i/p_i^{\max}$, and the second term reflects the owner's preference for charging soon after arrival. Both terms are convex, and for $\omega=\beta=0$ every EV is a pure price taker. The total cost in \eqref{eq:obj_site} is denoted by $J_{\mathrm{tot}}$.

\begin{assumption}[Feasibility]
\label{ass:feasible}
For every EV, $E_i\le p_i^{\max}|\Tset_i|\Delta t$, and there exist schedules that satisfy \eqref{eq:coupling_sum_site}--\eqref{eq:site_rating}.
\end{assumption}
Under Assumption~\ref{ass:feasible}, \eqref{eq:obj_site}--\eqref{eq:site_rating} is a feasible convex quadratic program (QP); hence, Karush--Kuhn--Tucker (KKT) points exist, and since $q(t)>0$, the optimal aggregate schedule $\mathbf{z}^\ast$ is unique, as are the individual schedules if $\beta>0$.

Each EV is represented by an \emph{EV-side agent}, i.e., the vehicle, its charger, the owner's application, or the manufacturer's cloud, which participates in a day-ahead coordination session $\Sset$. The agent need not be plugged in while the iterations run, because physical availability is captured by $\Tset_i$. The operator back end communicates with chargers through the Open Charge Point Protocol (OCPP)~\cite{OCPP201}, chargers communicate with vehicles through the interface of the International Organization for Standardization (ISO) standard 15118~\cite{ISO15118_20}, and application-based services route messages through manufacturer or aggregator clouds. Each hop is protected by mutually authenticated TLS; however, every intermediary terminates the session, reads the payload, and re-originates it; hence, TLS protects each link against outsiders but provides no end-to-end protection across intermediaries. Moreover, EVs have no P2P links, and every message to or from an EV passes through the operator.

\subsection{Coordination via Sharing-Form ADMM}
\label{subsec:distributed_optimization}
Problem \eqref{eq:obj_site}--\eqref{eq:site_rating} is an instance of the \emph{sharing problem}
\begin{equation}
\min_{\mathbf{x}_1,\ldots,\mathbf{x}_N}\ \sum_{i=1}^{N} h_i(\mathbf{x}_i)+g\Big(\sum_{i=1}^{N}\mathbf{x}_i\Big),
\label{eq:sharing_problem}
\end{equation}
with $h_i=\iota_{\mathcal{X}_i}+\varphi_i$ and $g=J+\iota_{\mathcal{Z}}$, where $\mathcal{X}_i$ is the set of schedules of EV $i$ that satisfy \eqref{eq:rate_window}--\eqref{eq:site_rating}, $\mathcal{Z}=\{\mathbf{z}\in\R^T:0\le z(t)\le P_{\mathrm{CS}}^{\max}\}$, and $\iota$ denotes an indicator function. Applying ADMM with the penalty $\rho>0$ and the scaled dual variable $\mathbf{u}$ to the equivalent two-block problem with copies $\mathbf{y}_i=\mathbf{x}_i$ and eliminating the copies yields~\cite[Sec.~7.3]{Boyd2011ADMM}
\begin{align}
\mathbf{x}_i^{k+1}&=\arg\min_{\mathbf{x}_i}\ h_i(\mathbf{x}_i)+\tfrac{\rho}{2}\big\|\mathbf{x}_i-\mathbf{x}_i^k+\bar{\mathbf{x}}^k-\bar{\mathbf{y}}^k+\mathbf{u}^k\big\|_2^2,\label{eq:sharing_x}\\
\bar{\mathbf{y}}^{k+1}&=\arg\min_{\bar{\mathbf{y}}}\ g(N\bar{\mathbf{y}})+\tfrac{N\rho}{2}\big\|\bar{\mathbf{y}}-\mathbf{u}^k-\bar{\mathbf{x}}^{k+1}\big\|_2^2,\label{eq:sharing_y}\\
\mathbf{u}^{k+1}&=\mathbf{u}^k+\bar{\mathbf{x}}^{k+1}-\bar{\mathbf{y}}^{k+1},\label{eq:sharing_u}
\end{align}
where bars denote averages over the $N$ EVs. Since each EV requires only the common vector $\bar{\mathbf{y}}^k-\bar{\mathbf{x}}^k-\mathbf{u}^k$, all EVs update in parallel, and since the scheme is two-block ADMM, it converges for any $\rho>0$ and any initialization~\cite[Thm.~8]{Eckstein1992DR}. With $\mathbf{S}^k=\sum_i\mathbf{x}_i^k=N\bar{\mathbf{x}}^k$, $\mathbf{z}^k=N\bar{\mathbf{y}}^k$, and the price $\boldsymbol{\lambda}^k=\rho\mathbf{u}^k$, the iteration proceeds as follows.

\noindent\textbf{Operator broadcast:}
At iteration $k$, the operator broadcasts the coordination vector $\mathbf{v}^k$ with components
\begin{equation}
{v^k(t)=\frac{1}{N}\big(z^k(t)-S^k(t)\big)-\frac{1}{\rho}\lambda^k(t),\quad t\in\Tset.}
\label{eq:v_def_unscaled}
\end{equation}
Each EV thus corrects only its $1/N$ share of the aggregate mismatch. The broadcast contains only aggregate information, and EVs do not exchange local profiles; hence, no EV can infer the schedule of another. The operator, however, observes the EV profiles in order to implement charging; privacy from the operator is therefore not an objective of this work.

\noindent\textbf{EV local update:}
By \eqref{eq:sharing_x}, each EV computes
\begin{align}
\mathbf{x}_i^{k+1}
&=\arg\min_{\mathbf{x}\in\mathcal{X}_i}\ \varphi_i(\mathbf{x})+\tfrac{\rho}{2}\big\|\mathbf{x}-\mathbf{x}_i^k-\mathbf{v}^k\big\|_2^2\notag\\
&=\Pi_{\mathcal{X}_i}\big(\mathbf{y}_i^k\big),\quad
\mathbf{y}_i^k=\frac{\rho(\mathbf{x}_i^k+\mathbf{v}^k)-\boldsymbol{\omega}_i+\beta\mathbf{x}_i^{\mathrm{p}}}{\rho+\beta},
\label{eq:ev_x_proj_v_unscaled}
\end{align}
where $\omega_i(t)=\omega(t-t_i^{\mathrm{a}})\Delta t$ on $\Tset_i$ and $\Pi_{\mathcal{X}_i}$ is the Euclidean projection onto $\mathcal{X}_i$. It has the water-filling form
\[
x_i^{k+1}(t)=\min\{p_i^{\max},\max\{0,y_i^k(t)-\pi_i\}\},\quad t\in\Tset_i,
\]
and $x_i^{k+1}(t)=0$ otherwise, where the scalar offset $\pi_i$ is obtained by bisection to satisfy \eqref{eq:site_rating}.

\noindent\textbf{Operator update:}
After receiving $\{\mathbf{x}_i^{k+1}\}$, the operator forms $\mathbf{S}^{k+1}$, solves \eqref{eq:sharing_y}, i.e., sets $z^{k+1}(t)=\Pi_{\mathcal{Z}}(\hat z^{k+1}(t))=\min\{P_{\mathrm{CS}}^{\max},\max\{0,\hat z^{k+1}(t)\}\}$, and updates the price:
\begin{align}
\hat z^{k+1}(t)&=\frac{{\frac{\rho}{N}}S^{k+1}(t)+\lambda^k(t)-\big(2q(t)D(t)+c(t)\big)\Delta t}{2q(t)\Delta t+{\rho/N}},
\label{eq:z_uncon_site_derived}\\
\lambda^{k+1}(t)&=\lambda^k(t)+{\tfrac{\rho}{N}}\big(S^{k+1}(t)-z^{k+1}(t)\big).
\label{eq:lambda_update_site}
\end{align}
The iteration terminates when the primal residual $r_\mathrm{p}^{k}=\|\mathbf{S}^k-\mathbf{z}^k\|_2/\sqrt{N}$ and the dual residual $r_\mathrm{d}^{k}=\rho(\sum_i\|\mathbf{e}_i^k\|_2^2)^{1/2}$, with $\mathbf{e}_i^k=\Delta\mathbf{x}_i^{k}-(\Delta\mathbf{S}^{k}-\Delta\mathbf{z}^{k})/N$ and $\Delta$ the change between consecutive iterations, fall below the tolerances $\varepsilon_p$ and $\varepsilon_d$~\cite[Sec.~3.3]{Boyd2011ADMM}. Because $r_\mathrm{d}$ contains the change of each individual schedule, termination certifies that the individual schedules, not merely their aggregate, have settled.

\begin{lemma}[Benign convergence]
\label{lem:benign}
Under Assumption~\ref{ass:feasible}, for any $\rho>0$ and any initialization, the iterates of \eqref{eq:v_def_unscaled}--\eqref{eq:lambda_update_site} converge to a primal--dual optimal point $(\mathbf{x}^\ast,\mathbf{z}^\ast,\boldsymbol{\lambda}^\ast)$ of \eqref{eq:obj_site}--\eqref{eq:site_rating}, and $r_\mathrm{p}^k,r_\mathrm{d}^k\to0$. Moreover, (a) each $\mathbf{x}_i^\ast$ minimizes $\boldsymbol{\lambda}^{\ast\top}\mathbf{x}+\varphi_i(\mathbf{x})$ over $\mathcal{X}_i$, i.e., at convergence every EV responds optimally to the price, given its preferences; and (b) $\lambda^\ast(t)=[2q(t)(D(t)+z^\ast(t))+c(t)]\Delta t$ whenever $0<z^\ast(t)<P_{\mathrm{CS}}^{\max}$, i.e., the price equals the marginal grid cost.
\end{lemma}
Lemma~\ref{lem:benign} holds only if every participant follows \eqref{eq:v_def_unscaled}--\eqref{eq:lambda_update_site}. The next section examines what happens otherwise.

\section{Threat Model}
\label{sec:threat_model}

The scheme of Section~\ref{subsec:distributed_optimization} assumes that every EV receives the same coordination signal $\mathbf{v}^k$ and computes its update \eqref{eq:ev_x_proj_v_unscaled}. This section models adversaries that violate these assumptions from either side and characterizes their attacks. We consider a computationally bounded adversary, formally a probabilistic polynomial-time (PPT) adversary, that is \emph{static}, i.e., it fixes its targets before the session starts. It may take the following three forms, which may also collude:

\begin{itemize}\setlength{\itemsep}{1pt}\setlength{\topsep}{2pt}
    \item \emph{A1, biased operator:} The operator itself is malicious. It sends different signals to different EVs, or withholds them, over valid TLS sessions and observes all uploads. Because every message passes through it, it can also delay, drop, or selectively deliver the messages of the key generation (Section~\ref{subsec:setup}), and it may collude with up to $f$ signers. Its objective is to shift payments among EVs while keeping aggregate indicators close to normal. Falsifying the signal identically for all EVs, in contrast, cannot bias the charging costs of some EVs relative to others and is outside the scope of this paper.
    \item \emph{A2, MITM attacker:} The operator is honest, but the attacker controls a node on the communication links of a set $\Aset$ of EVs, such as a compromised gateway. On these paths, it can read, modify, drop, delay, replay, or inject messages, but it cannot access other hops; it does not break TLS but resides at an endpoint of a TLS hop. Its objective is to increase the victims' costs or to disrupt the coordination.
    \item \emph{A3, Byzantine EVs:} The attacker controls the software of some EVs, for example, through compromised vehicle applications or manufacturer clouds. A Byzantine EV may upload arbitrary schedules, withhold uploads, misreport its constraints, and, if it is a signer, sign arbitrary messages, in which case it counts toward $f$. Its objective is to shift costs onto honest EVs~\cite{Fard2023ISIE,FakouriFard2025Targeted} or to disrupt convergence.
\end{itemize}
No adversary can forge the signatures of honest parties, break TLS on hops that it does not control, or find collisions of the hash function, except with negligible probability. A Byzantine EV may also misreport $E_i$, $\Tset_i$, or $p_i^{\max}$; it then behaves as an honest EV with these requirements and must charge and pay accordingly, and data-driven detectors~\cite{Shafee2020LyingEV} can complement the proposed defense against such misreports.

To formalize the downlink attacks, let $\mathcal{A}$ be the set of attacked EVs, $A=|\mathcal{A}|$, and let $a_i(t)\in\{-1,0,+1\}$ be a deterministic attack pattern for EV $i\in\Aset$, where $+1$ and $-1$ mark the slots toward and away from which its charging is shifted. Each attacked EV $i$ is also assigned a direction $\gamma_i\in\{-1,0,+1\}$ and a magnitude $\tau_i\ge0$. The injected bias is
\begin{equation}
\delta_i(t)=\gamma_i\tau_i a_i(t),\quad i\in\mathcal{A},\ t\in\Tset.
\label{eq:delta_i_def}
\end{equation}
Thus, an attacked EV receives $\tilde v_i^k(t)=v^k(t)+\delta_i(t)$ for $k\ge k_0$, where $k_0$ is the attack onset, and $\tilde v_i^k(t)=v^k(t)$ otherwise. A benign EV computes \eqref{eq:ev_x_proj_v_unscaled}, whereas an attacked EV without protection computes
\begin{equation}
\mathbf{x}_i^{k+1}
=
\arg\min_{\mathbf{x}\in\mathcal{X}_i}\ \varphi_i(\mathbf{x})+\tfrac{\rho}{2}\big\|\mathbf{x}-\mathbf{x}_i^k-\tilde{\mathbf{v}}_i^k\big\|_2^2,
\quad i\in\mathcal{A}.
\label{eq:ev_update_v_only_attacked}
\end{equation}
A Byzantine EV that adds $\boldsymbol{\delta}_i$ to its own update follows the same recursion; hence, the following result also covers A3 and includes the participant attacks of~\cite{Fard2023ISIE,FakouriFard2025Targeted} as a special case.

\begin{proposition}[Convergence under undefended attacks]
\label{prop:stealth}
Suppose Assumption~\ref{ass:feasible} holds and the bias \eqref{eq:delta_i_def} is applied for all $k\ge k_0$ without protection. Then, from iteration $k_0$ on, \eqref{eq:v_def_unscaled}--\eqref{eq:lambda_update_site} with \eqref{eq:ev_update_v_only_attacked} is exactly sharing-form ADMM applied to
\begin{equation}
\min_{\{\mathbf{x}_i\},\mathbf{z}}\ J_{\mathrm{tot}}-\rho\sum_{i\in\Aset}\boldsymbol{\delta}_i^{\top}\mathbf{x}_i
\quad\text{s.t. \eqref{eq:coupling_sum_site}--\eqref{eq:site_rating}}.
\label{eq:perturbed_problem}
\end{equation}
Consequently, for any onset $k_0$: (a) the iterates converge and $r_\mathrm{p}^k,r_\mathrm{d}^k\to0$; (b) at the limit $(\tilde{\mathbf{x}},\tilde{\mathbf{z}},\tilde{\boldsymbol{\lambda}})$, each attacked EV's schedule minimizes $(\tilde{\boldsymbol{\lambda}}-\rho\boldsymbol{\delta}_i)^{\top}\mathbf{x}+\varphi_i(\mathbf{x})$ over $\mathcal{X}_i$, i.e., the victim responds to the falsified price $\tilde{\boldsymbol{\lambda}}-\rho\boldsymbol{\delta}_i$, whereas every other EV responds to $\tilde{\boldsymbol{\lambda}}$; and (c) the attack outcome is an optimal solution of the convex QP \eqref{eq:perturbed_problem}.
\end{proposition}

Proposition~\ref{prop:stealth} implies that monitoring the residuals or the objective cannot reveal the attack, because the attacked iteration converges normally, albeit to an incorrect operating point. It also explains how the adversaries choose the sign of the bias. Under A1, the operator sets $\gamma_i=+1$ for \emph{disadvantaged} EVs, so that expensive slots appear cheap to them, and $\gamma_i=-1$ for \emph{advantaged} EVs, so that cheap slots appear even cheaper; the two groups exchange slots, which redistributes payments while the aggregate load, and hence $J$, remains close to nominal. Under A2, the MITM attacker sets $\gamma_i=+1$ for all victims. The resulting damage and the parameters that maximize it can be characterized in closed form.

\begin{corollary}[Damage of an undefended attack]
\label{cor:damage}
Under the assumptions of Proposition~\ref{prop:stealth}, let $\mathrm{sp}_{\Tset_i}(\mathbf{w})=\max_{t\in\Tset_i}w(t)-\min_{t\in\Tset_i}w(t)$ denote the span of a vector $\mathbf{w}$ over the window of EV $i$, let $\tilde C_i(\mathbf{x})=\tilde{\boldsymbol{\lambda}}^{\top}\mathbf{x}+\varphi_i(\mathbf{x})$ be its local cost at the price that the other EVs face, and let $\bar{\mathbf{x}}_i$ minimize $\tilde C_i$ over $\mathcal{X}_i$, i.e., be EV $i$'s best response to that price. Then every $i\in\Aset$ with $\gamma_i\neq0$ satisfies
\begin{equation}
\tilde C_i(\tilde{\mathbf{x}}_i)-\tilde C_i(\bar{\mathbf{x}}_i)\le\frac{E_i}{\Delta t}\,\rho\tau_i\,\mathrm{sp}_{\Tset_i}(\mathbf{a}_i),
\label{eq:damage}
\end{equation}
and the extra payment $\tilde{\boldsymbol{\lambda}}^{\top}(\tilde{\mathbf{x}}_i-\bar{\mathbf{x}}_i)$ never exceeds $(E_i/\Delta t)\,\mathrm{sp}_{\Tset_i}(\tilde{\boldsymbol{\lambda}})$. Moreover, if $\gamma_i=+1$, $a_i(t)=+1$ on a set of the most expensive slots of $\Tset_i$, $a_i(t)=-1$ on the other slots of $\Tset_i$, and $\tau_i>\mathrm{sp}_{\Tset_i}(\tilde{\boldsymbol{\lambda}})/(2\rho)$, then the falsified price ranks every slot with $a_i(t)=+1$ below every slot with $a_i(t)=-1$, and a price-taking victim fills the most expensive slots first.
\end{corollary}

Corollary~\ref{cor:damage} shows that the attack parameters require no inner optimization. The damage increases linearly with $\tau_i$ until the falsified price reorders the victim's slots and is bounded by the price span thereafter. The attacker therefore sets $\tau_i$ to a multiple of the threshold $\mathrm{sp}_{\Tset_i}(\tilde{\boldsymbol{\lambda}})/(2\rho)$, which A1 computes directly and A2 infers from the intercepted broadcasts, since $\mathbf{v}^k\approx-\boldsymbol{\lambda}^k/\rho$ near convergence; preference costs dampen the response of the victim, which requires a larger multiple. Because the damage scales with $E_i$ and vanishes for a pattern that is constant on $\Tset_i$, the attacker tailors each pattern $\mathbf{a}_i$ to the window of EV $i$ and targets EVs whose windows contain both cheap and expensive hours. A1 observes all $E_i$ and $\Tset_i$ and chooses its targets optimally. A grid-aware operator places the $+1$ slots of disadvantaged EVs only in expensive hours with spare grid capacity, which leaves the peak load unchanged and the attack inconspicuous. These findings motivate the following defense.

\section{Proposed Secure Distributed EV Charging Framework}
\label{sec:proposed_secure_framework}

\begin{figure}[!t]
\centering
\includegraphics[width=\linewidth]{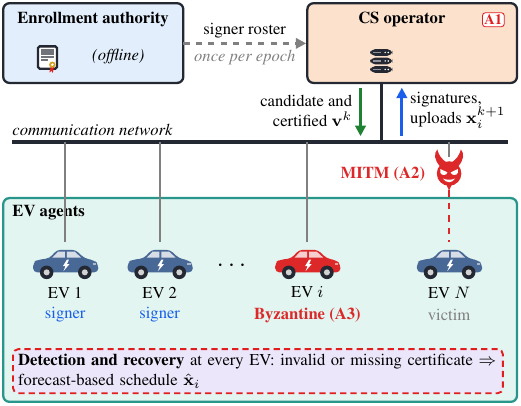}
\caption{Framework architecture and adversary positions.}
\label{fig:architecture}
\end{figure}

\begin{figure}[!t]
\centering
\includegraphics[width=\linewidth]{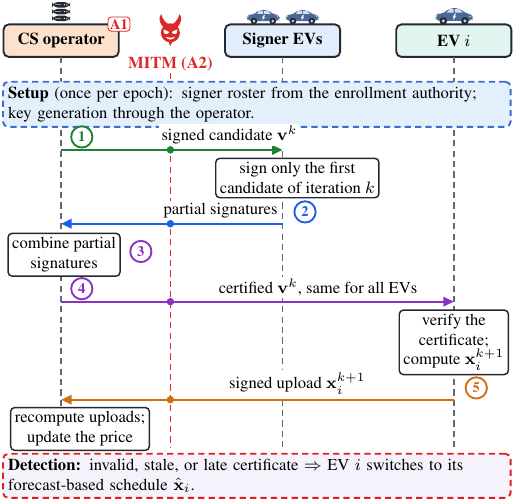}
\caption{Message flow of one certified iteration.}
\label{fig:protocol}
\end{figure}

As argued in Section~I, neither TLS nor a signature by the operator can reveal A1 or A2. The proposed framework therefore protects both directions of every iteration. On the downlink, each broadcast must carry a certificate, i.e., a threshold signature that a sufficiently large group of designated signers must produce jointly, which detects A1 and A2. On the uplink, the operator recomputes every upload, which exposes Byzantine EVs. Since detection cannot prevent an adversary from altering or withholding a signal, it is complemented by an ML-based recovery that keeps the victim in the optimization. Fig.~\ref{fig:architecture} shows the resulting architecture, which involves only the cyber layer. The CS operator, which may be biased (A1), exchanges all messages with the EV agents over a communication network, in which a MITM attacker may control the link to some EVs (A2), and some EV agents may be Byzantine (A3). A subset of the EVs acts as signers; a committee of institutions can take this role instead (Section~\ref{subsec:setup}). On the downlink, the operator sends candidate and certified signals $\mathbf{v}^k$; on the uplink, it receives partial signatures and the uploads $\mathbf{x}_i^{k+1}$. An EV that receives an invalid or missing certificate detects the attack and switches to a forecast-based schedule $\hat{\mathbf{x}}_i$ (Section~\ref{subsec:recovery}). Fig.~\ref{fig:protocol} details the message flow of one iteration (Section~\ref{subsec:certified}), which Section~\ref{subsec:security} analyzes.

\subsection{Enrollment and Key Generation}
\label{subsec:setup}
The certificates use threshold Boneh--Lynn--Shacham (BLS) signatures. Let $\Gone$, $\Gtwo$, and $\mathbb{G}_T$ be cyclic groups of prime order $l$ with a bilinear pairing $e:\Gone\times\Gtwo\rightarrow\mathbb{G}_T$, a generator $G\in\Gtwo$, and a hash-to-curve function $H:\{0,1\}^\ast\rightarrow\Gone$. A BLS key pair is $(\mathrm{sk},\mathrm{pk})$ with $\mathrm{pk}=\mathrm{sk}\,G$, and a signature $\sigma=\mathrm{sk}\,H(m)$ on message $m$ is valid if
\begin{equation}
e(\sigma,G)=e(H(m),\mathrm{pk}).
\label{eq:sig_check}
\end{equation}
In threshold BLS, signer $j$ holds a share $\mathrm{sk}_j$ of $\mathrm{sk}$ and issues the partial signature $\sigma_j=\mathrm{sk}_j H(m)$, verifiable under its public share $\mathrm{pk}_j=\mathrm{sk}_j G$; any $P$ signers combine their partial signatures by Lagrange interpolation into a $\sigma$ that satisfies \eqref{eq:sig_check}, whereas fewer than $P$ signers cannot. Threshold BLS is existentially unforgeable under adaptive chosen-message attacks (EUF-CMA), i.e., no computationally bounded adversary can produce a valid signature on a message that the signers have not signed, as long as fewer than $P$ signers are corrupted; this holds in the random oracle model (ROM), which treats the hash function as an ideal random function~\cite{BLS2004,Boldyreva2003Threshold}. Security is stated with respect to a security parameter $\eta$, to which the bit length of $l$ is proportional (the BLS12-381 curve targets approximately 128-bit security), and $\negl(\eta)$ denotes a function that vanishes faster than the inverse of any polynomial in $\eta$.

Since every message passes through the operator, EVs that share no information fixed in advance could be partitioned by a biased operator into groups that run separate key generations and sessions, each of which appears complete to its members, and no protocol could detect such a partition. We therefore adopt the weakest common reference that excludes this attack, namely the \emph{membership} of the signer set. Because this anchor is static and carries no session data, it cannot deliver the coordination signal, in contrast to a bulletin board.

Each signer holds a credential, i.e., a certificate that binds a signing key to its identity, issued by an enrollment authority (EA) that the operator does not control; the V2G public-key infrastructure that issues ISO~15118 contract certificates and TLS certificates can serve as the EA~\cite{ISO15118_20}. Once per epoch, e.g., one month, the offline EA signs a \emph{roster} that lists the station, the epoch, the credentials of the $N_s$ signers, and the parameters $(f,d,P)$; it takes no part in key generation or sessions. Because the operator can neither issue credentials nor alter the roster, it cannot inflate the signer set with fictitious identities, i.e., mount a Sybil attack; preventing such attacks requires a logically centralized identity authority~\cite{Douceur2002Sybil}, a role that the EA fulfills without being online. Identities that the adversary does control count toward $f$.

The signers are either the EV-side agents of the EVs enrolled in the station's program, which can sign on days on which their EVs do not charge, or a committee of $M$ independent parties, e.g., the distribution system operator, an aggregator, and an auditor, in which case EVs only verify certificates.

We assume that at most $f$ signers are Byzantine, i.e., they may collude with the adversary and sign anything, and at most $d$ honest signers are unavailable in any iteration, i.e., offline, delayed past the deadline, or isolated by A2. The EA sets the signing threshold such that
\begin{equation}
\frac{N_s+f}{2}<P\le N_s-f-d,
\label{eq:threshold}
\end{equation}
which is feasible if and only if $N_s\ge3f+2d+1$. The lower bound guarantees consistency, and the upper bound guarantees that certificates continue to form (Theorem~\ref{thm:byz_collusion}). With $d=0$ and $N_s=3f+1$, \eqref{eq:threshold} yields the classical requirement that $P=2f+1$ signers, i.e., two thirds, must sign, and a positive $d$ trades tolerance of Byzantine signers for tolerance of missing signatures; for example, $N_s=50$ signers tolerate $(f,d)=(10,9)$ with $P=31$, and a committee of $M=4$ institutions tolerates $f=1$ with $P=3$. The condition concerns only the signers. EVs that are not signers, join late, or go offline during a session affect neither safety nor liveness, and Byzantine EVs count toward $f$ only if they are signers.

Once per epoch, the signers generate the threshold key using the distributed key generation (DKG) protocol of Gennaro~\emph{et al.}~\cite{Gennaro2007DKG}. Each signer $j$ shares a random secret $\psi_j(0)$ with a random degree-$(P-1)$ polynomial $\psi_j$ and publishes Pedersen commitments to its coefficients; signers verify their shares and complain if a check fails, and dealers that cannot answer complaints are disqualified, leaving the qualified set $\mathcal{Q}$. Only then does each qualified dealer publish $\psi_{j,h}G$ for its coefficients, from which every signer obtains $\mathrm{pk}=\sum_{j\in\mathcal{Q}}\psi_j(0)G$ and the public shares, and signer $u$ sets $\mathrm{sk}_u=\sum_{j\in\mathcal{Q}}\psi_j(u)$. The original protocol assumes a broadcast channel among the signers, which is unavailable in our setting; we replace it as follows: 1) every message that a signer would broadcast is signed with its credential and forwarded by the operator, and every private share is encrypted under the recipient's credential key; the operator can therefore delay or drop messages but neither alter nor read them; 2) each phase ends with a \emph{certified transcript}, i.e., the operator forwards the collected messages to all signers, and a signer signs the digest with its credential, at most once per phase, only if the transcript contains every message it sent and, for the sharing phase, only if $|\mathcal{Q}|\ge f+1$; and 3) a signer accepts a phase only if at least $P$ roster members signed the same digest. The argument that protects every iteration, namely that any two groups of $P$ signers share at least one honest signer, also protects the key generation.

\begin{lemma}[Agreement of the key generation]
\label{lem:dkg}
Let $P>(N_s+f)/2$ and the credential signatures be EUF-CMA secure. Then, except with negligible probability, (a) at most one transcript per phase can be certified, and all honest signers that complete it agree on $\mathcal{Q}$ and $\mathrm{pk}$; (b) $\mathcal{Q}$ contains at least one honest dealer; and (c) if the operator forwards all messages and at most $f+d$ signers do not respond, every phase is certified.
\end{lemma}

A signer whose shares are inconsistent with the accepted transcript cannot produce valid partial signatures and therefore counts toward $d$. The key generation requires $O(N_s^2)$ forwarded messages once per epoch, which favors a committee for large fleets, and an EV that is not a signer can verify $\mathrm{pk}$ from the certified transcripts and the roster.

The recovery of Section~\ref{subsec:recovery} uses a learned price forecaster $\mathcal{M}$. If the forecaster or its inputs could differ across EVs, the operator could steer the recovery itself, and the ML component would thus open a new attack surface. Once per epoch, the signers therefore certify the digest of $\mathcal{M}$ under $\mathrm{pk}$, after checking its error on certified past sessions against a level set by the EA, and every EV verifies this certificate. At the start of each day-ahead session $\Sset$, the operator obtains a certificate on the message $m^0$ that contains the forecast inputs of \eqref{eq:ml_lambda_ev}, so that all EVs compute the same forecast $\hat{\boldsymbol{\lambda}}$. Each EV initializes at $\mathbf{x}_i^0=\Pi_{\mathcal{X}_i}(\bar{\mathbf{u}}_i-\hat{\boldsymbol{\lambda}}/\tilde\rho)$, where $\bar{\mathbf{u}}_i$ spreads $E_i$ uniformly over $\Tset_i$ and $\tilde\rho$ is the recovery parameter of Section~\ref{subsec:recovery}, and the operator sets $\boldsymbol{\lambda}^0=\hat{\boldsymbol{\lambda}}$ and $\mathbf{z}^0=\mathbf{S}^0$; by Lemma~\ref{lem:benign}, this initialization does not change the benign limit but gives every EV a price-aware reference schedule.

\subsection{Certified Iteration, Detection, and Recovery}
\label{subsec:certified}
At iteration $k$ of session $\Sset$, the plain broadcast of \eqref{eq:v_def_unscaled} is replaced by the five steps shown in Fig.~\ref{fig:protocol}. 1) The operator computes $\mathbf{v}^k$, forms the message $m^k=(\Sset,k,\mathbf{v}^k)$, and sends the candidate $m^k$, signed with its own key, to the signers. 2) An honest signer $j$ verifies the operator's signature and $(\Sset,k)$ and returns the partial signature $\sigma_j^k=\mathrm{sk}_j H(m^k)$ for the first valid candidate of $(\Sset,k)$ only. 3) The operator combines any $P$ partial signatures into $\sigma^k$ and, in an \emph{optimistic verification}, checks only the combined signature with \eqref{eq:sig_check}, verifying partial signatures individually only if that check fails. 4) The operator broadcasts $(m^k,\sigma^k)$, and EV $i$ accepts $\mathbf{v}^k$ only if $\sigma^k$ verifies under $\mathrm{pk}$, the embedded $(\Sset,k)$ is current, and the message arrives before the deadline $T_{\mathrm{dl}}$; otherwise, it raises a detection event (Section~\ref{subsec:recovery}). 5) EV $i$ uploads $(\mathbf{x}_i^{k+1},\Sset,k{+}1)$ signed with its credential. The operator discards uploads with invalid signatures or stale identifiers and \emph{recomputes} every other upload from the EV's declared constraints $\mathcal{X}_i$, its last accepted upload, $\mathbf{v}^k$, and the certified forecast, checking, up to numerical tolerance, that it equals the prescribed update \eqref{eq:ev_x_proj_v_unscaled} or, after a detection event, the prescribed recovery \eqref{eq:recovery}. An EV whose upload is missing or matches neither is \emph{frozen} at its last accepted upload for the rest of the session.

Steps 1--4 instantiate the threshold-certified broadcast of~\cite{Cachin2001Broadcast}, with the session and iteration identifiers providing the standard protection against replay; the recomputation in step 5, the key generation of Section~\ref{subsec:setup}, and the recovery described below are specific to the proposed framework. Since honest signers sign at most one candidate per $(\Sset,k)$, two different broadcasts cannot both be certified (Theorem~\ref{thm:byz_collusion}). Since uploads are signed end to end, A2 can drop but not alter them, and since they are recomputed, at the cost of one projection per EV, a Byzantine EV cannot deviate without being frozen. An EV that holds a valid operator signature on a broadcast that conflicts with the certified one possesses verifiable evidence of an inconsistency attack, whereas an invalid operator signature indicates tampering on the path. We assume synchronous rounds, i.e., honest messages arrive within a known bound below $T_{\mathrm{dl}}$, and treat late messages as missing. No step requires P2P links or any channel other than the operator, and, unlike in a ledger, nothing is replicated or stored. Per iteration, plain ADMM exchanges $2N$ messages, and certification adds $2N$ with EV signers or $2M$ with a committee of $M$ parties; hence, communication remains linear in $N$, and every EV verifies one constant-size certificate, whatever the fleet size.

\label{subsec:recovery}
EV $i$ raises a detection event at the first iteration $k_i^0$ in which the broadcast fails Step~4. Every falsified or withheld broadcast triggers such an event in the iteration in which it occurs (Theorem~\ref{thm:byz_collusion}), regardless of how $\boldsymbol{\delta}_i$ is constructed or how plausible it appears, and attack-free operation triggers none (Corollary~\ref{cor:detection}). Detection alone, however, leaves the victim without a coordination signal, and if many EVs merely ceased updating, the attack would turn from deceptive into disruptive. The defense thus includes a forecast-based recovery.

For this purpose, we train an offline regression model $\mathcal{M}$ that predicts the 24-hour price vector from inputs that are available before the session and certified at setup:
\begin{equation}
\hat{\lambda}(t)=\mathcal{M}\!\big(t,\,E,\,\hat D(t),\,c(t),\,\boldsymbol{\chi}(t)\big),\quad t\in\Tset,
\label{eq:ml_lambda_ev}
\end{equation}
where $E=\sum_iE_i$ is the total energy request and $\hat D(t)$ is the day-ahead forecast of the base load. The fleet features $\boldsymbol{\chi}(t)$ summarize the registrations that the EVs submit at the start of the session: the number of EVs available in slot $t$, their aggregate rated power $\sum_{i:t\in\Tset_i}p_i^{\max}$, the average power that they require, $\sum_{i:t\in\Tset_i}E_i/(|\Tset_i|\Delta t)$, both power values relative to $P_{\mathrm{CS}}^{\max}$, and $P_{\mathrm{CS}}^{\max}$ itself. These features indicate the slots in which the demand approaches the station rating, where congestion raises the price above the marginal grid cost; such day-specific price peaks cannot be inferred from the tariff or from past prices alone. The training labels are the benign prices of past days, taken as the marginal cost $[2q(t)(D(t)+z^\ast(t))+c(t)]\Delta t$, which equals $\lambda^\ast(t)$ wherever $0<z^\ast(t)<P_{\mathrm{CS}}^{\max}$ (Lemma~\ref{lem:benign}(b)), keeps the congestion premium of $\lambda^\ast(t)$ where $z^\ast(t)=P_{\mathrm{CS}}^{\max}$, and fixes the price in slots without charging, where any value below the marginal cost is optimal. In deployment, the certified broadcasts of past sessions provide every EV with the same labels.

At convergence, $\mathbf{S}^\ast=\mathbf{z}^\ast$, and \eqref{eq:v_def_unscaled} gives $\mathbf{v}^\ast=-\boldsymbol{\lambda}^\ast/\rho$; that is, a price forecast reconstructs the coordination signal that the attacker falsified or withheld. At its detection event, EV $i$ therefore substitutes $\hat{\mathbf{v}}=-\hat{\boldsymbol{\lambda}}/\tilde\rho$ for the rejected broadcast in \eqref{eq:ev_x_proj_v_unscaled}, applied to its last accepted schedule $\mathbf{r}_i=\mathbf{x}_i^{k_i^0}$:
\begin{equation}
\hat{\mathbf{x}}_i=\arg\min_{\mathbf{x}\in\mathcal{X}_i}\ \varphi_i(\mathbf{x})+\hat{\boldsymbol{\lambda}}^{\top}\mathbf{x}+\tfrac{\tilde\rho}{2}\|\mathbf{x}-\mathbf{r}_i\|_2^2,
\label{eq:recovery}
\end{equation}
where $\tilde\rho>0$ is a recovery parameter. The EV then retains $\hat{\mathbf{x}}_i$, uploads it in every iteration, and remains in recovery for the rest of the session. The rule is the EV's optimal response to the forecast, anchored at the last schedule coordinated with the fleet. A small $\tilde\rho$ relies on the forecast, whereas a large $\tilde\rho$ keeps the EV close to $\mathbf{r}_i$. Applying the rule once, rather than iterating with the reconstructed signal, which is accurate only near convergence, keeps it valid regardless of when the attack starts (Theorems~\ref{thm:conv_recovery}--\ref{thm:baselines}). Moreover, because an EV that enters recovery remains in it for the rest of the session, an adversary that alternately delivers and withholds certificates cannot make an EV switch modes indefinitely. The historical average or the last certified price can replace $\hat{\boldsymbol{\lambda}}$, as compared in Section~\ref{sec:implementation_results}.

Because every $\mathbf{x}\in\mathcal{X}_i$ delivers the same energy, adding a constant to $\hat{\boldsymbol{\lambda}}$ does not change \eqref{eq:recovery}, and Theorem~\ref{thm:cost_recovery} turns this structure into a learning objective. The cost of recovery depends on the forecast only through the span of its error within each window, and a forecast with an offset in level but a correct shape therefore incurs no cost. We therefore train $\mathcal{M}$ on day-centered targets $\lambda^\ast(t)-\frac{1}{T}\sum_{s}\lambda^\ast(s)$ and select it by the span of its error rather than by the mean absolute error (MAE) or the mean absolute percentage error (MAPE).

\subsection{Theoretical Guarantees}
\label{subsec:security}

We first establish the security of the certified broadcast against a PPT adversary that controls the CS operator and the network and corrupts a set $\mathcal{C}$ of signers with $|\mathcal{C}|\le f$, including any Byzantine EV signer. Let $\epsilon_{\mathrm{BLS}}$ and $\epsilon_{\mathrm{DL}}$ be the maximum probabilities that a PPT adversary forges a threshold BLS signature and computes a discrete logarithm in $\Gtwo$, respectively; both are negligible in $\eta$. In the ROM, an adversary with at most $q_H$ oracle queries finds a hash collision, i.e., $m\neq m'$ with $H(m)=H(m')$, with probability at most $\epsilon_H=q_H^2/(2l)$.

\begin{theorem}[Consistency, liveness, and key privacy]
\label{thm:byz_collusion}
Assume that threshold BLS is EUF-CMA secure in the ROM when fewer than $P$ signers are corrupted and that the key was generated as in Section~\ref{subsec:setup}. Then: (i) \emph{Safety:} If $P>(N_s+f)/2$, the probability that an honest EV accepts a certified message that fewer than $P-|\mathcal{C}|$ honest signers signed, or that two honest EVs accept different certified messages for the same pair $(\Sset,k)$, satisfies
\begin{equation}
\Pr[\text{consistency failure}]\le2\epsilon_{\mathrm{BLS}}+\epsilon_H.
\label{eq:sec_bound_A}
\end{equation}

\noindent(ii) \emph{Liveness:} If the operator follows the protocol, at most $f$ signers withhold their partial signatures, and at most $d$ honest signers are unavailable, then a certificate forms in every iteration whenever $P\le N_s-f-d$.

\noindent(iii) \emph{Key privacy:} The probability that the adversary outputs the group secret key $\mathrm{sk}$ satisfies
\begin{equation}
\Pr[\text{key recovery}]\le\epsilon_{\mathrm{DL}}+\negl(\eta).
\label{eq:sec_bound_B}
\end{equation}
\end{theorem}

The threshold in (i) cannot be lowered. If $P\le(N_s+f)/2$, then $2(P-f)\le N_s-f$, and an adversary that corrupts $f$ signers can therefore send two different messages for the same $(\Sset,k)$ to disjoint groups of $P-f$ honest signers and obtain a certificate on each. By (ii), missing signatures can delay certification, which triggers recovery, but never cause an EV to accept an inconsistent broadcast. Parts (i) and (ii) adapt the classical argument for certified broadcast~\cite{Cachin2001Broadcast} and part (iii) that of~\cite{Gennaro2007DKG}; the new element is that they hold for a key generated through the untrusted operator (Lemma~\ref{lem:dkg}), without a trusted dealer or a broadcast channel among the signers.

Replay protection follows, as in standard certified broadcast, from binding each certificate to the session and iteration identifiers. Honest signers and EVs check these identifiers before signing or accepting; hence, a replayed or altered broadcast passes verification only through a new threshold BLS forgery or a hash collision, and
\begin{equation}
\Pr[\text{replay accepted}]\le\epsilon_{\mathrm{BLS}}+\epsilon_H.
\label{eq:sec_bound_replay}
\end{equation}

\begin{corollary}[Detection without false alarms]
\label{cor:detection}
Under the assumptions of Theorem~\ref{thm:byz_collusion}, (a) if no adversary is active, at most $d$ honest signers are unavailable, and all messages meet the deadline, no honest EV raises a detection event, and the defended iteration coincides with the benign one; (b) a broadcast that differs from the certified one, is replayed, or is withheld raises a detection event at every honest EV that receives or awaits it, except with the probability bounded in \eqref{eq:sec_bound_A} and \eqref{eq:sec_bound_replay}; and (c) if the operator follows the protocol, every upload that deviates from the prescribed update or recovery is detected in the iteration in which it arrives, whereas the uploads of honest EVs always pass.
\end{corollary}
Parts (a) and (b) follow from Theorem~\ref{thm:byz_collusion}, \eqref{eq:sec_bound_replay}, and the correctness of threshold BLS, and part (c) holds because both prescribed updates are deterministic functions of data held by the operator. In contrast to a learned detector, neither check trades missed detections against false alarms. It remains to determine how the EVs that detect an attack should proceed.

\label{subsec:recovery_analysis}
We now analyze the defended iteration after detection. Let $\Aset$ be the set of EVs that stop coordinating, at iterations $k_i^0\le\bar k<\infty$, either because they detect an attack and recover by \eqref{eq:recovery} or because the operator freezes them; let $\hat{\mathbf{x}}_i\in\mathcal{X}_i$ be the recovered or frozen schedule; and let $\mathcal{P}_\Aset$ denote problem \eqref{eq:obj_site}--\eqref{eq:site_rating} with $\mathbf{x}_i$ fixed to $\hat{\mathbf{x}}_i$ for every $i\in\Aset$.

\begin{theorem}[Convergence and feasibility under recovery]
\label{thm:conv_recovery}
Suppose Assumption~\ref{ass:feasible} holds and the certified broadcast is computed by \eqref{eq:v_def_unscaled} from the accepted uploads. If $\mathcal{P}_\Aset$ is feasible, then for any detection iterations and any state at detection, the defended iteration converges to a primal--dual optimal point of $\mathcal{P}_\Aset$, and $r_\mathrm{p}^k,r_\mathrm{d}^k\to0$. Moreover, $\mathcal{P}_\Aset$ is feasible whenever
\begin{equation}
\sum_{i\in\Aset}\big(\hat x_i(t)-x_i^\ast(t)\big)\le P_{\mathrm{CS}}^{\max}-z^\ast(t),\quad t\in\Tset,
\label{eq:feasibility}
\end{equation}
where $(\mathbf{x}^\ast,\mathbf{z}^\ast)$ is the benign optimum of Lemma~\ref{lem:benign}; \eqref{eq:feasibility} holds for any recovered schedules if $\sum_{i\in\Aset}p_i^{\max}\le\min_t\big(P_{\mathrm{CS}}^{\max}-z^\ast(t)\big)$ or $\sum_{i=1}^Np_i^{\max}\le P_{\mathrm{CS}}^{\max}$. If $\mathcal{P}_\Aset$ is infeasible, $r_\mathrm{p}^k$ stays bounded away from zero, and the operator observes it.
\end{theorem}

\begin{theorem}[Cost of recovery]
\label{thm:cost_recovery}
Let $(\mathbf{x}^\ast,\mathbf{z}^\ast,\boldsymbol{\lambda}^\ast)$ be the benign optimal point of Lemma~\ref{lem:benign}, let $C_i^\ast(\mathbf{x})=\boldsymbol{\lambda}^{\ast\top}\mathbf{x}+\varphi_i(\mathbf{x})$ be EV $i$'s local cost at the benign price, and let $\epsilon_i=\mathrm{sp}_{\Tset_i}(\hat{\boldsymbol{\lambda}}-\boldsymbol{\lambda}^\ast)$ be the span of the forecast error. Then $\hat{\mathbf{x}}_i$ is invariant to adding a constant to $\hat{\boldsymbol{\lambda}}$, and every $i\in\Aset$ recovering by \eqref{eq:recovery} satisfies
\begin{equation}
C_i^\ast(\hat{\mathbf{x}}_i)-C_i^\ast(\mathbf{x}_i^\ast)
\le
\frac{E_i}{\Delta t}\,\epsilon_i
+\frac{\tilde\rho}{2}\big\|\mathbf{x}_i^\ast-\mathbf{r}_i\big\|_2^2,
\label{eq:cost_bound_ev}
\end{equation}
whereas every schedule $\mathbf{x}\in\mathcal{X}_i$ satisfies
\begin{equation}
\boldsymbol{\lambda}^{\ast\top}(\mathbf{x}-\mathbf{x}_i^\ast)\le\frac{E_i}{\Delta t}\,\mathrm{sp}_{\Tset_i}(\boldsymbol{\lambda}^\ast).
\label{eq:envelope}
\end{equation}
If, in addition, \eqref{eq:feasibility} holds and $0<z^\ast(t)<P_{\mathrm{CS}}^{\max}$ for all $t$, the optimal total cost $J_\Aset$ of $\mathcal{P}_\Aset$ satisfies
\begin{align}
J_\Aset-J_{\mathrm{tot}}^\ast
&\le
\sum_{i\in\Aset}\Big[\frac{E_i}{\Delta t}\,\epsilon_i+\frac{\tilde\rho}{2}\big\|\mathbf{x}_i^\ast-\mathbf{r}_i\big\|_2^2\Big]
\notag\\
&\quad
+\Delta t\,\max_{t}q(t)\,\Big\|\sum_{i\in\Aset}(\hat{\mathbf{x}}_i-\mathbf{x}_i^\ast)\Big\|_2^2,
\label{eq:cost_bound_sys}
\end{align}
where $J_{\mathrm{tot}}^\ast$ is the benign optimal total cost and, for a frozen EV, the bracket is replaced by $\frac{E_i}{\Delta t}\mathrm{sp}_{\Tset_i}(\boldsymbol{\lambda}^\ast)+\varphi_i(\hat{\mathbf{x}}_i)-\varphi_i(\mathbf{x}_i^\ast)$.
\end{theorem}

To benchmark the recovery rule, let $L_i^{\mathbf{w}}(\mathbf{x})=\mathbf{w}^{\top}\mathbf{x}+\varphi_i(\mathbf{x})-\min_{\mathbf{x}'\in\mathcal{X}_i}[\mathbf{w}^{\top}\mathbf{x}'+\varphi_i(\mathbf{x}')]$ be the extra local cost of $\mathbf{x}\in\mathcal{X}_i$ at prices $\mathbf{w}$, so that, by Lemma~\ref{lem:benign}(a), the left-hand side of \eqref{eq:cost_bound_ev} is $L_i^{\boldsymbol{\lambda}^\ast}(\hat{\mathbf{x}}_i)$, and let $\|\cdot\|_{\Tset_i}$ be the Euclidean norm over $\Tset_i$.

\begin{theorem}[Recovery versus baselines]
\label{thm:baselines}
For every $i\in\Aset$ that recovers by \eqref{eq:recovery}:

\noindent(a) \emph{Versus detection only:} the schedule $\mathbf{r}_i$ that a detection-only defense would hold satisfies
\begin{equation}
L_i^{\boldsymbol{\lambda}^\ast}(\hat{\mathbf{x}}_i)\le L_i^{\boldsymbol{\lambda}^\ast}(\mathbf{r}_i)+\min\Big\{\frac{E_i}{\Delta t}\epsilon_i,\ \frac{|\Tset_i|\,\epsilon_i^2}{16\tilde\rho}\Big\}.
\label{eq:vs_freeze}
\end{equation}

\noindent(b) \emph{Lower bounds for price takers ($\varphi_i\equiv0$):} Suppose $\Tset_i$ splits into two sets, in each of which EV $i$ can deliver $E_i$. For every $S>0$ and every price-agnostic rule, e.g., uniform charging or charging on arrival, there are prices $\mathbf{w}$ with $\mathrm{sp}_{\Tset_i}(\mathbf{w})=S$ and $L_i^{\mathbf{w}}\ge E_iS/(2\Delta t)$. For every $\epsilon>0$ and every rule that knows the prices only through a forecast $\hat{\mathbf{w}}$, there are prices $\mathbf{w}$ with $\mathrm{sp}_{\Tset_i}(\mathbf{w}-\hat{\mathbf{w}})=\epsilon$ and $L_i^{\mathbf{w}}\ge E_i\epsilon/(2\Delta t)$, whereas a price-taking response to $\hat{\mathbf{w}}$ guarantees $L_i^{\mathbf{w}}\le E_i\epsilon/\Delta t$.

\noindent(c) \emph{Stability:} for any two forecasts, $\|\hat{\mathbf{x}}_i(\hat{\boldsymbol{\lambda}})-\hat{\mathbf{x}}_i(\hat{\boldsymbol{\lambda}}')\|_2\le\min_{\vartheta\in\R}\|\hat{\boldsymbol{\lambda}}-\hat{\boldsymbol{\lambda}}'-\vartheta\mathbf{1}\|_{\Tset_i}/(\tilde\rho+\beta)$.
\end{theorem}

\begin{corollary}[Reduction of every attack to an erasure]
\label{cor:content}
Suppose the assumptions of Theorem~\ref{thm:byz_collusion} hold, the forecaster and the session inputs are certified, and the certified broadcast is computed by \eqref{eq:v_def_unscaled} from uploads that pass the recomputation, as under A2 and A3; under A1, this excludes the consistent falsification discussed in Section~\ref{sec:threat_model}. Then, except with negligible probability, an adversary can influence the honest EVs only through the \emph{erasure pattern}, i.e., which EVs stop coordinating and when, and through the constraints that Byzantine EVs declare. Two strategies that induce the same erasure pattern and declarations lead to identical schedules for all honest EVs, whatever content they falsify. Hence, Theorems~\ref{thm:conv_recovery}--\ref{thm:baselines} hold uniformly over all adversary strategies, whereas without the defense, the adversary sets each victim's effective price, and only the worst-case bound \eqref{eq:envelope} limits its damage (Corollary~\ref{cor:damage}).
\end{corollary}

Condition \eqref{eq:feasibility} compares, in every slot, the extra load of the EVs in recovery with the headroom of the benign schedule; it holds for any attack on EVs whose rates fit within the minimum headroom, regardless of the onset, and Theorem~\ref{thm:baselines}(c) shows that the recovered schedules, and hence their extra load in each slot, change continuously with the forecast. A forecast that misses the congestion hours of the day therefore places the fixed load of the recovered EVs in slots without headroom, which can violate \eqref{eq:feasibility}; Section~\ref{sec:implementation_results} confirms this effect. In \eqref{eq:cost_bound_ev}, the first term, the price of relying on a forecast, depends only on the \emph{shape} of the forecast error within the window, which defines the training target of Section~\ref{subsec:recovery}; compared with the worst-case bound \eqref{eq:envelope}, which an undefended attacker can reach (Corollary~\ref{cor:damage}), the span of the prices is replaced by the span of the forecast error, up to an anchor term that reflects the attack timing. The last term of \eqref{eq:cost_bound_sys} is the coordination lost because EVs in recovery no longer respond to the fleet.

Theorem~\ref{thm:baselines} compares the rule with the two defenses that it integrates. A purely cryptographic defense can only retain $\mathbf{r}_i$ or revert to a price-agnostic schedule, which by (b) can lose half of the worst-case bound, whereas by (a), the proposed rule is never worse than retaining $\mathbf{r}_i$ by more than \eqref{eq:vs_freeze}, a margin that vanishes as $\tilde\rho$ increases. A purely ML-based defense, i.e., price taking on the forecast, guarantees $E_i\epsilon_i/\Delta t$, which \eqref{eq:recovery} attains as $\tilde\rho\to0$, and by (b), no rule that knows the prices only through a forecast can guarantee less than half of this value. The rule is thus minimax-optimal up to a factor of two against early attacks and, for a large $\tilde\rho$, approaches the detection-only schedule, which is near-optimal against late attacks; by (c), it degrades gracefully with the forecast error.

\section{Simulation Results}
\label{sec:implementation_results}

\subsection{Setup, Data, and Metrics}
\label{subsec:setup_results}
The daily instances were constructed from the EVWatts Public Database~\cite{pritchard2026evwatts}, which contains anonymized charging sessions and trips of EVs in the United States. We selected 500 vehicles with frequent charging in 2022 and formed one instance per day, which yields 365 daily fleets with 25 to 163 EVs and 108 EVs on average. For each EV, the window $\Tset_i$ extends from plug-in to the next departure, $E_i$ is the energy delivered in the session, and $p_i^{\max}$ follows from the recorded charging power; the fleet therefore contains both Level~2 and direct-current fast charging, as recorded. The base load $D(t)$ is the hourly system load of the California Independent System Operator~\cite{EnergyOnlineCAISO} of the same day of the year, scaled to the size of a distribution bus, and $c(t)$ is the business EV time-of-use tariff of Pacific Gas and Electric (PG\&E)~\cite{PGE_BEV_Tariff}, whose peak price applies from 16:00 to 21:00. The station rating $P_{\mathrm{CS}}^{\max}$ is set 20\% above the peak load that results when every EV charges uniformly over its window, which keeps the problem feasible and lets congestion occur on busy days. The experiments compare uncoordinated charging, coordinated charging without attacks, and coordinated charging under attack with and without the defense. The model uses $T=24$, $\Delta t=1$~h, $q(t)=2\times10^{-6}$~\$/(kW$^2$h), $\omega=0.03$~\$/kWh per slot, $\beta=0.02$, $\rho=0.003N$, and $\varepsilon_p=\varepsilon_d=10^{-4}$. The framework was implemented in Python with compiled EV updates, and threshold BLS uses the BLS12-381 curve. Benign ADMM converged on all 365 days with a median of 116 iterations.

We report the grid cost $J(\mathbf{z})$ and the total cost $J_{\mathrm{tot}}$ of \eqref{eq:obj_site} only as changes relative to benign operation, because both are system-level indices rather than cash flows. $J$ adds to the tariff cost of the EV energy a grid-stress cost of the entire bus load, including the non-EV load that no EV pays for, and $J_{\mathrm{tot}}$ adds the preference costs of the EVs. Their absolute values therefore exceed the EV payments without implying a loss for the station, whose profit is the EV payments minus the tariff cost of the delivered energy, and differences of $J_{\mathrm{tot}}$ are in dollar-equivalent terms per day. The \emph{load-flattening score} (LFS) measures how the EV schedule changes the shape of the bus load. With $\mathrm{PAR}(\mathbf{y})=\max_ty(t)/\big(\frac{1}{T}\sum_ty(t)\big)$ denoting the peak-to-average ratio of a load profile $\mathbf{y}$, we define
\begin{equation}
\mathrm{LFS}=100\%\times\frac{\mathrm{PAR}(\mathbf{D})-\mathrm{PAR}(\mathbf{D}+\mathbf{z})}{\mathrm{PAR}(\mathbf{D})},
\label{eq:lfs}
\end{equation}
i.e., the relative reduction of the peak-to-average ratio achieved by the EV schedule; positive values indicate that charging fills valleys, and negative values indicate that it accentuates peaks. The daily payment of EV $i$ is $\sum_t\lambda(t)x_i(t)\Delta t$ at the final iterate, and its annual cost scales its average price on the day to a typical annual charging energy of 3,854~kWh. The station profit is $\sum_t(\lambda(t)-c(t))z(t)\Delta t$, i.e., the payments of the EVs minus the cost of the energy at the tariff.

\subsection{Price Forecaster}
\label{subsec:ml_results}
The forecaster was trained on 288 days and tested on 77 held-out days, with entire days held out, so that no hour of a test day was used in training. Nine regression models were compared, namely Ridge, Lasso, ElasticNet, decision tree, support vector regression (SVR), random forest, gradient boosting, XGBoost, and LightGBM~\cite{Ke2017LightGBM}. Table~\ref{tab:ml_models} reports their coefficient of determination ($R^2$), MAE, and MAPE on the training and test sets, together with the energy-weighted error span, i.e., the average of $\epsilon_i$ over the windows of the test EVs weighted by $E_i$, which governs the cost of recovery by Theorem~\ref{thm:cost_recovery}. The table also lists two references without learning, namely the tariff and the hourly average of the training prices.

\begin{table}[!t]
\centering
\renewcommand{\arraystretch}{1.05}
\scriptsize
\setlength{\tabcolsep}{3.2pt}
\caption{Training and Test Performance of the Price Forecasters}
\label{tab:ml_models}
\begin{tabular}{lcccccc}
\toprule
\multirow{2}{*}{\textbf{Model}} & \multicolumn{2}{c}{\textbf{Training}} & \multicolumn{4}{c}{\textbf{Testing}} \\
\cmidrule(lr){2-3}\cmidrule(lr){4-7}
& $R^2$ & MAE & $R^2$ & MAE & MAPE (\%) & Span (\$/kWh) \\
\midrule
LightGBM         & 0.9964 & 0.0047 & 0.9695 & \textbf{0.0139} & \textbf{6.20} & \textbf{0.0550} \\
Random forest    & 0.9926 & 0.0055 & 0.9666 & 0.0140 & 6.32 & 0.0552 \\
XGBoost          & 0.9959 & 0.0054 & 0.9678 & 0.0145 & 6.59 & 0.0570 \\
SVR              & 0.9638 & 0.0135 & \textbf{0.9699} & 0.0150 & 6.60 & 0.0581 \\
Gradient boosting& 0.9907 & 0.0083 & 0.9683 & 0.0151 & 6.81 & 0.0589 \\
Decision tree    & 0.9797 & 0.0099 & 0.9459 & 0.0186 & 8.41 & 0.0751 \\
Ridge            & 0.5823 & 0.0689 & 0.5786 & 0.0689 & 29.74 & 0.1977 \\
Lasso            & 0.5823 & 0.0690 & 0.5787 & 0.0689 & 29.74 & 0.1977 \\
ElasticNet       & 0.5823 & 0.0690 & 0.5786 & 0.0689 & 29.74 & 0.1977 \\
\midrule
Tariff only      & -- & -- & 0.3203 & 0.0683 & 24.65 & 0.1881 \\
Historical average & -- & -- & 0.6757 & 0.0618 & 27.60 & 0.1333 \\
\bottomrule
\end{tabular}
\end{table}

The three linear models reach a test $R^2$ of only 0.58, because the benign price contains congestion premiums that appear only when the available demand approaches the station rating. Such threshold effects cannot be represented by a linear function of the inputs, whereas tree ensembles capture them and reach test $R^2$ values between 0.967 and 0.970. LightGBM attains the lowest test MAE, the lowest test MAPE, and the smallest error span of all models. It also generalizes best over time and reaches an $R^2$ of 0.878 when trained on earlier days and tested on later ones, compared with 0.844 to 0.876 for the other nonlinear models. SVR has a marginally higher test $R^2$, but its error span is 6\% larger, and by Theorem~\ref{thm:cost_recovery} the span, not $R^2$, determines the cost of recovery. LightGBM was therefore selected for the proposed recovery. Its error span of 0.055~\$/kWh is 59\% smaller than that of the historical average and 71\% smaller than that of the tariff, because the fleet features $\boldsymbol{\chi}(t)$ reveal the congestion hours of each day, which neither reference can anticipate.

\subsection{Coordination Benefit, Attack Impact, and Defense}
\label{subsec:attack_results}
Coordination is needed on every day of the data. Without coordination, each EV charges at its rated power from a random start time within its availability window. The windows, energies, and rated powers are those of the data, and only the start time is random; both cases therefore serve the same demand. Over 200 random draws on each of the 365 days, coordination lowers the EV energy cost on every day, by 10.4\% in the median, reduces the share of EV energy in the peak-price hours from 23.9\% to 12.8\%, and lowers the peak load on 74.2\% of the days, whereas 16.0\% of the uncoordinated draws exceed the station rating and coordination never does.
\begin{figure}[!t]
\centering
\subfloat[\label{fig:unc_total}]{\includegraphics[height=\dimexpr(\linewidth-7pt)*1000/2558\relax]{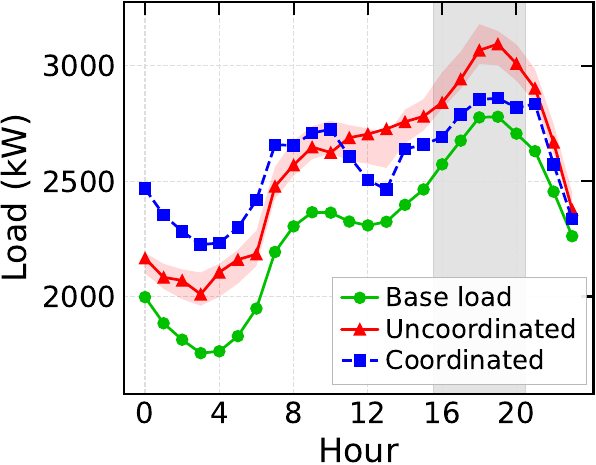}}\hspace{6pt}%
\subfloat[\label{fig:unc_ev}]{\includegraphics[height=\dimexpr(\linewidth-7pt)*1000/2558\relax]{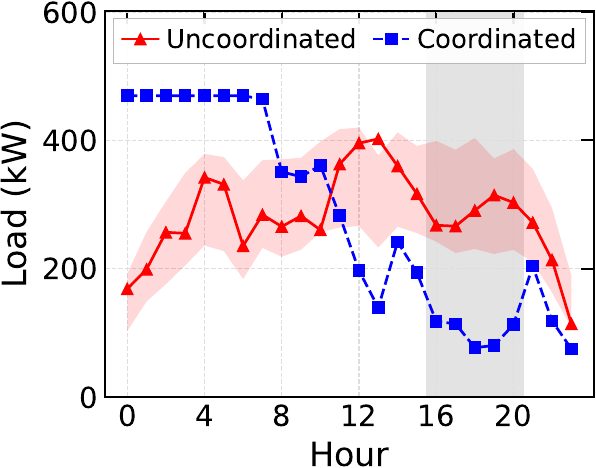}}
\caption{Uncoordinated versus coordinated charging on a high-demand day: (a) total bus load and (b) EV load. The red band covers 5\%--95\% of 200 random uncoordinated draws, and gray marks the peak-price hours.}
\label{fig:uncoordinated}
\end{figure}
The benefit grows with the charging demand. On the tenth of the days with the most charged energy, coordination lowers the EV energy cost by 15.9\% and the peak load by 2.1\% in the median, compared with 7.5\% and 0.6\% on the tenth with the least energy. Fig.~\ref{fig:uncoordinated} illustrates these effects on a high-demand day with 161 EVs and 6.8~MWh of charging. Uncoordinated charging stacks the EV load on the evening peak of the base load and raises the peak bus load to 3,095~kW, whereas coordination moves the charging into the night valley and lowers this peak by 7.6\% to 2,859~kW. Coordination thereby raises the LFS from 0.9\% to 8.4\%, reduces the share of EV energy in the peak-price hours from 22.2\% to 7.4\%, and lowers the EV energy cost by 15.3\%, while every EV still receives its full energy. The coordinated schedule is the ADMM schedule of Section~\ref{subsec:distributed_optimization}, which raises the EV load to the station rating of 470~kW in the night valley, where the base load leaves the most headroom; the peak of the EV load alone therefore does not measure the grid impact of charging, whereas the peak of the bus load does. Coordination does not flatten the load completely, because ADMM minimizes the grid cost together with the preference costs of the EVs, which favor charging soon after arrival, and the station rating caps the night charging.

The attack case study considers a held-out day with 142 EVs, on which the base load peaks within the peak-price hours and benign coordination creates no new load peak.
\begin{figure*}[!b]
\centering
\subfloat[\label{fig:A1_load}]{\includegraphics[height=\dimexpr(\linewidth-13pt)*1000/4065\relax]{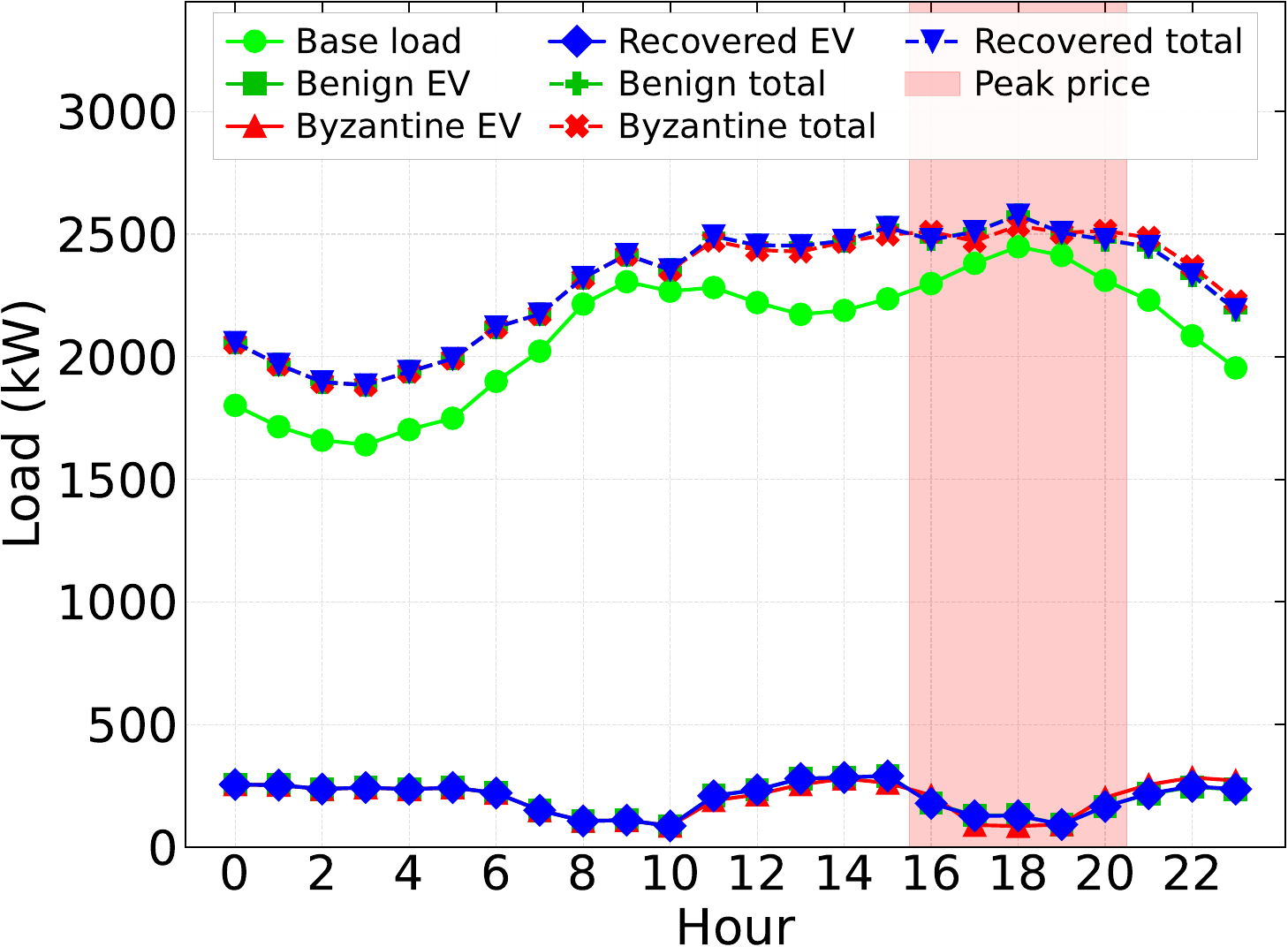}}\hspace{6pt}%
\subfloat[\label{fig:A2_load}]{\includegraphics[height=\dimexpr(\linewidth-13pt)*1000/4065\relax]{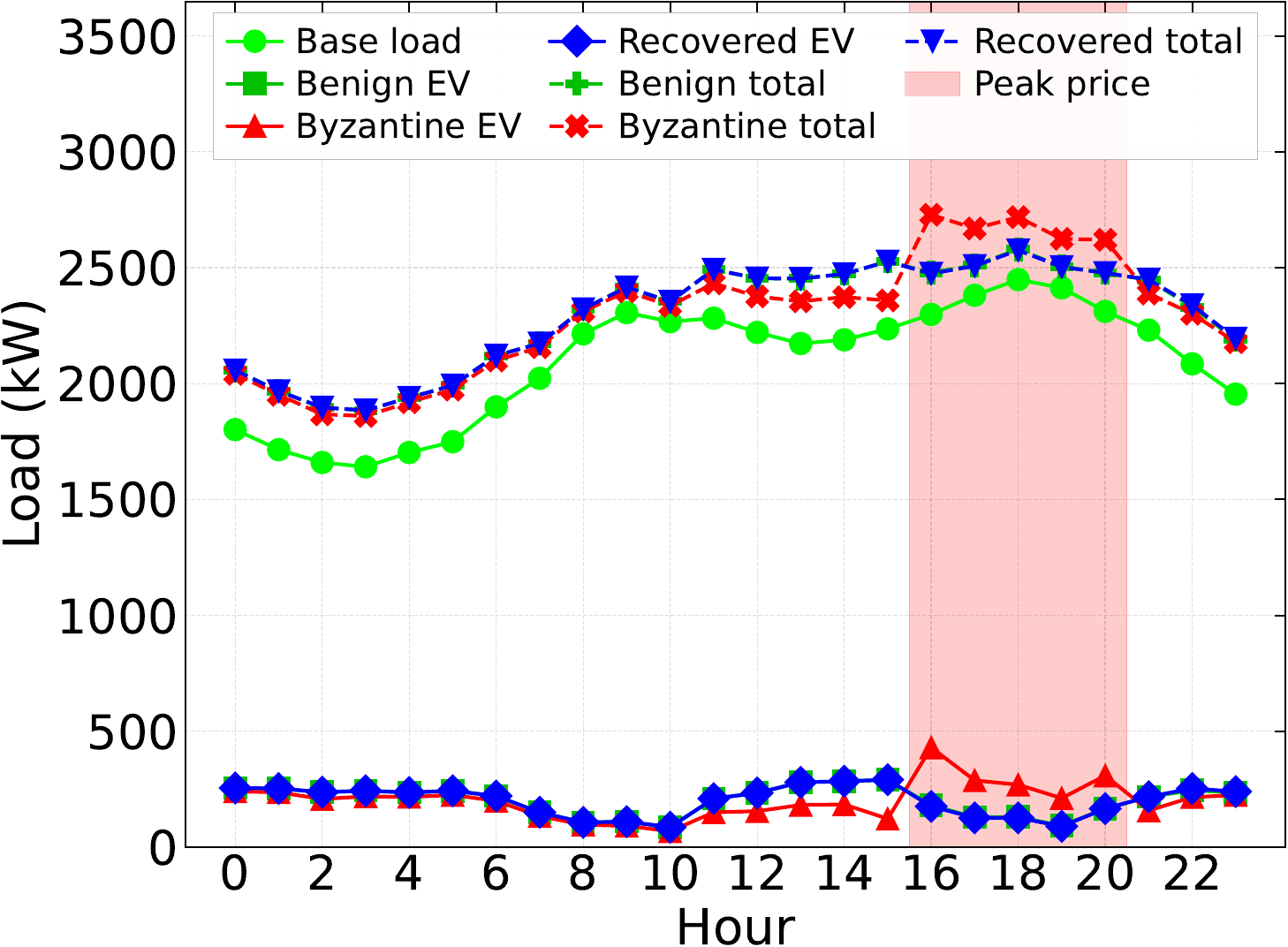}}\hspace{6pt}%
\subfloat[\label{fig:price_both}]{\includegraphics[height=\dimexpr(\linewidth-13pt)*1000/4065\relax]{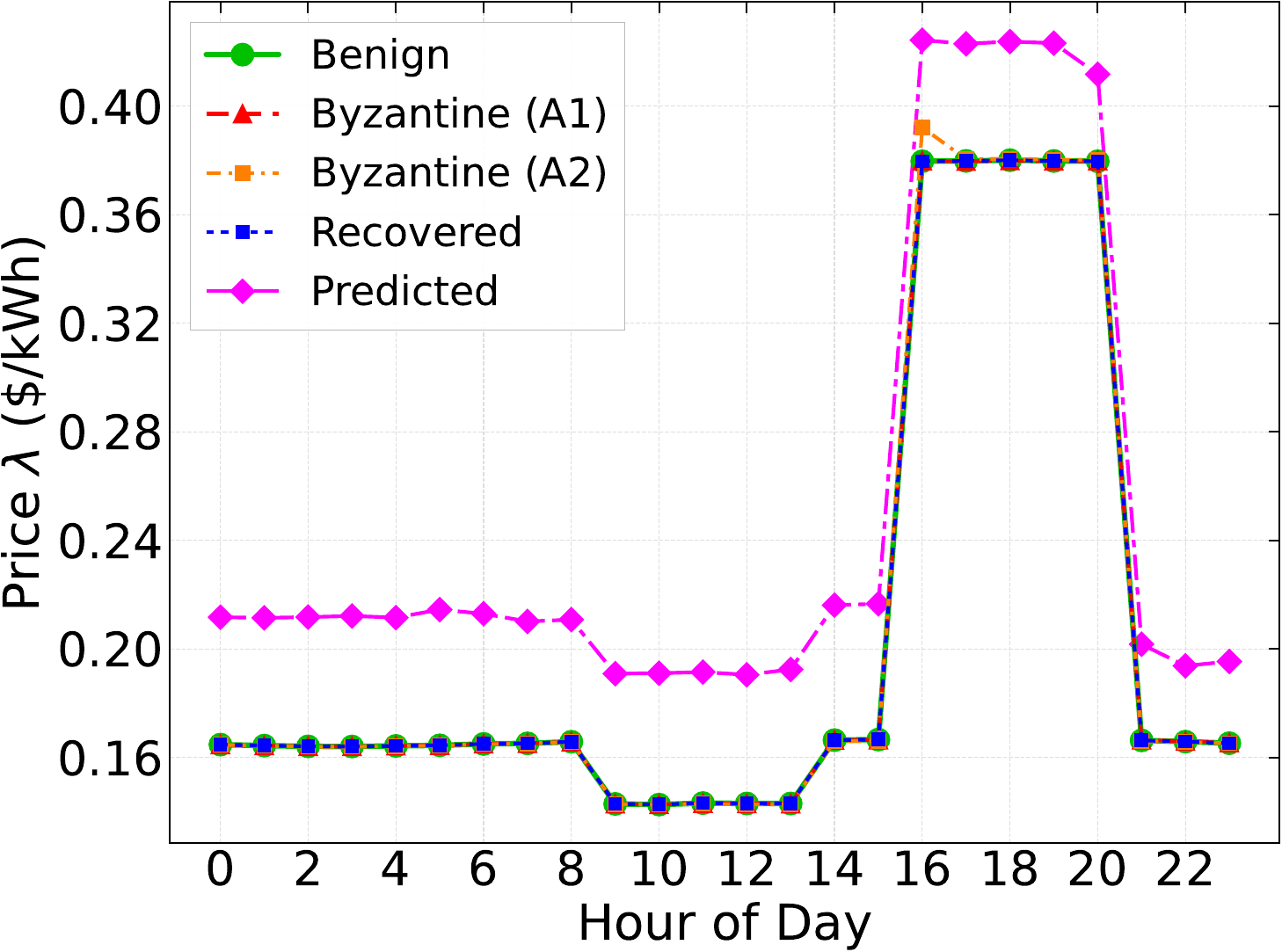}}
\caption{Load and price profiles: (a) load under A1, (b) load under A2, and (c) final prices under A1 and A2. The shaded hours carry the peak price.}
\label{fig:load_price_both}
\vspace{8pt}
\end{figure*}
This day differs from that of Fig.~\ref{fig:uncoordinated}, because the attack analysis requires a day without a new benign load peak, which isolates the load changes caused by the attacks. Both attacks start at iteration $k_0=30$ and follow Corollary~\ref{cor:damage}, with patterns $\mathbf{a}_i$ tailored to the window of each victim. A1 targets 10 EVs, namely 5 disadvantaged EVs ($\gamma_i=+1$), whose $+1$ slots are expensive hours with spare grid capacity, and 5 advantaged EVs ($\gamma_i=-1$), whose cheapest slots are made to appear even cheaper. A2 targets the 40 EVs whose payments it can raise most and sets $a_i(t)=+1$ in the peak-price slots of each window. In both attacks, $\tau_i$ is 20 to 50 times the threshold of Corollary~\ref{cor:damage}, and the defense uses $\tilde\rho=0.005$.

Fig.~\ref{fig:load_price_both} shows the load and the final price for benign operation, under attack (labeled Byzantine), and with the defense (labeled Recovered), together with the forecast (labeled Predicted). Benign coordination shifts charging out of the peak-price hours, where the base load also peaks. Under A1, the total load follows its benign profile throughout the day, and its peak even decreases by 1.7\%, because the disadvantaged EVs move only into expensive hours with spare capacity, while the advantaged EVs leave these hours. The aggregate indicators therefore reveal nothing, as Proposition~\ref{prop:stealth} predicts. Under A2, the victims are pushed into the peak-price hours, which raises the total load in these hours and its peak by 5.8\%. The final prices under both attacks, shown together in Fig.~\ref{fig:load_price_both}(c), nearly coincide with the benign price, because the tariff dominates them. A load shift of 250~kW changes the grid-stress term by only about \$0.001/kWh, and only the congestion at 16:00 under A2 raises the price visibly, by \$0.012/kWh. The attacks therefore leave no trace in the published price, and their damage appears in the schedules and payments. In both cases, the recovered load and price coincide with their benign profiles. The forecast deviates from the benign price by an almost constant offset, which does not affect recovery, because only the shape of the forecast error matters (Theorem~\ref{thm:cost_recovery}), not its level.

\begin{figure*}[!t]
\centering
\subfloat[\label{fig:A1_annual}]{\includegraphics[height=\dimexpr(0.71\linewidth-7pt)*1000/3527\relax]{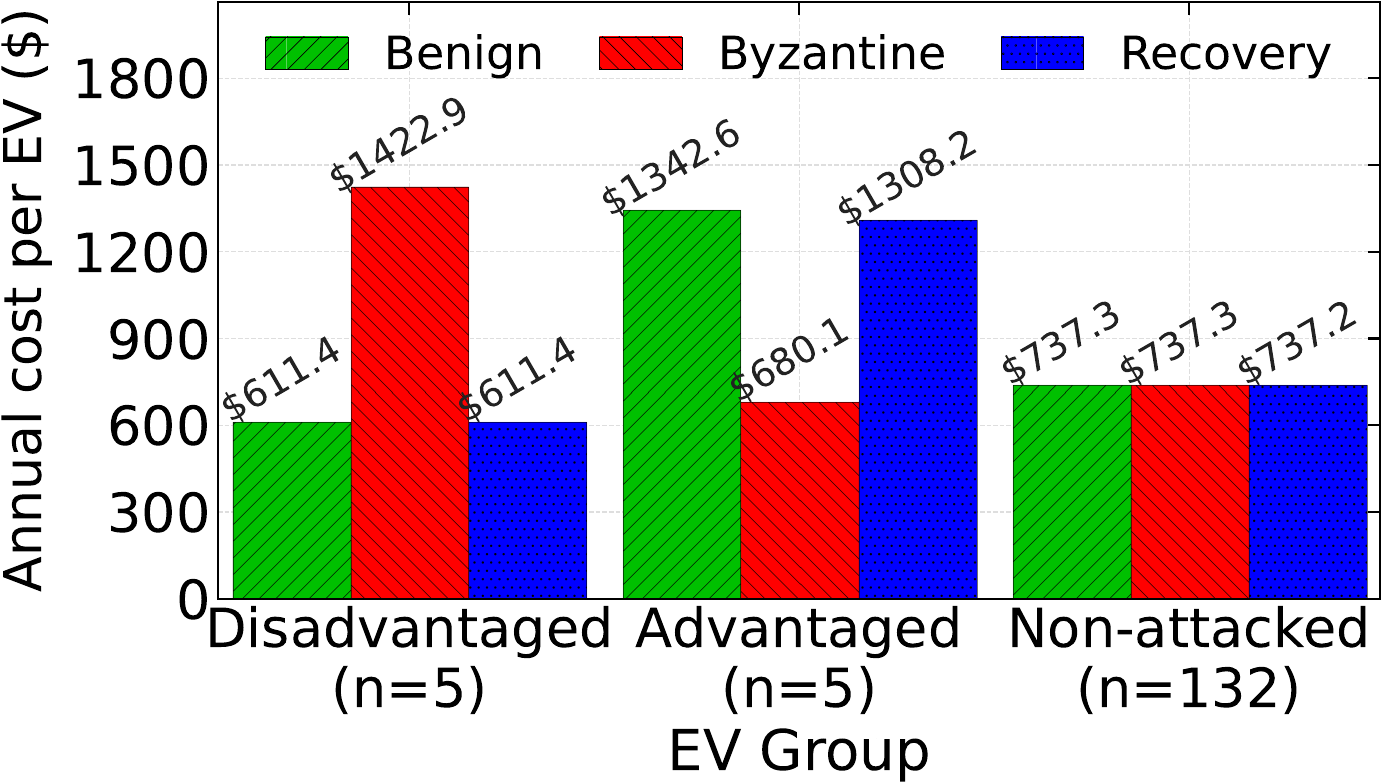}}\hspace{6pt}%
\subfloat[\label{fig:A2_annual}]{\includegraphics[height=\dimexpr(0.71\linewidth-7pt)*1000/3527\relax]{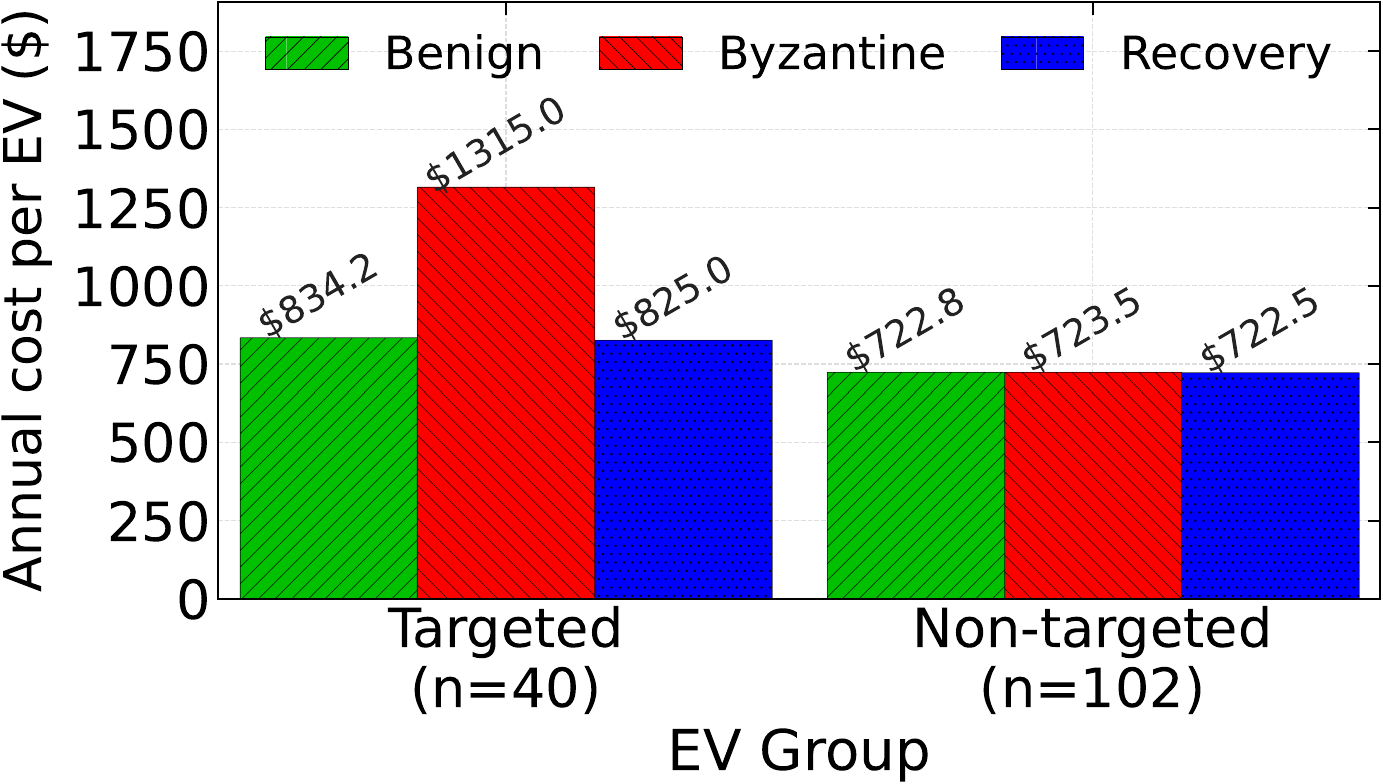}}
\caption{Annual charging cost per EV under (a) A1 and (b) A2.}
\label{fig:annual_cost_comparison}
\end{figure*}

\begin{figure*}[!t]
\centering
\subfloat[\label{fig:A1_ben}]{\includegraphics[height=\dimexpr(\linewidth-13pt)*1000/6190\relax]{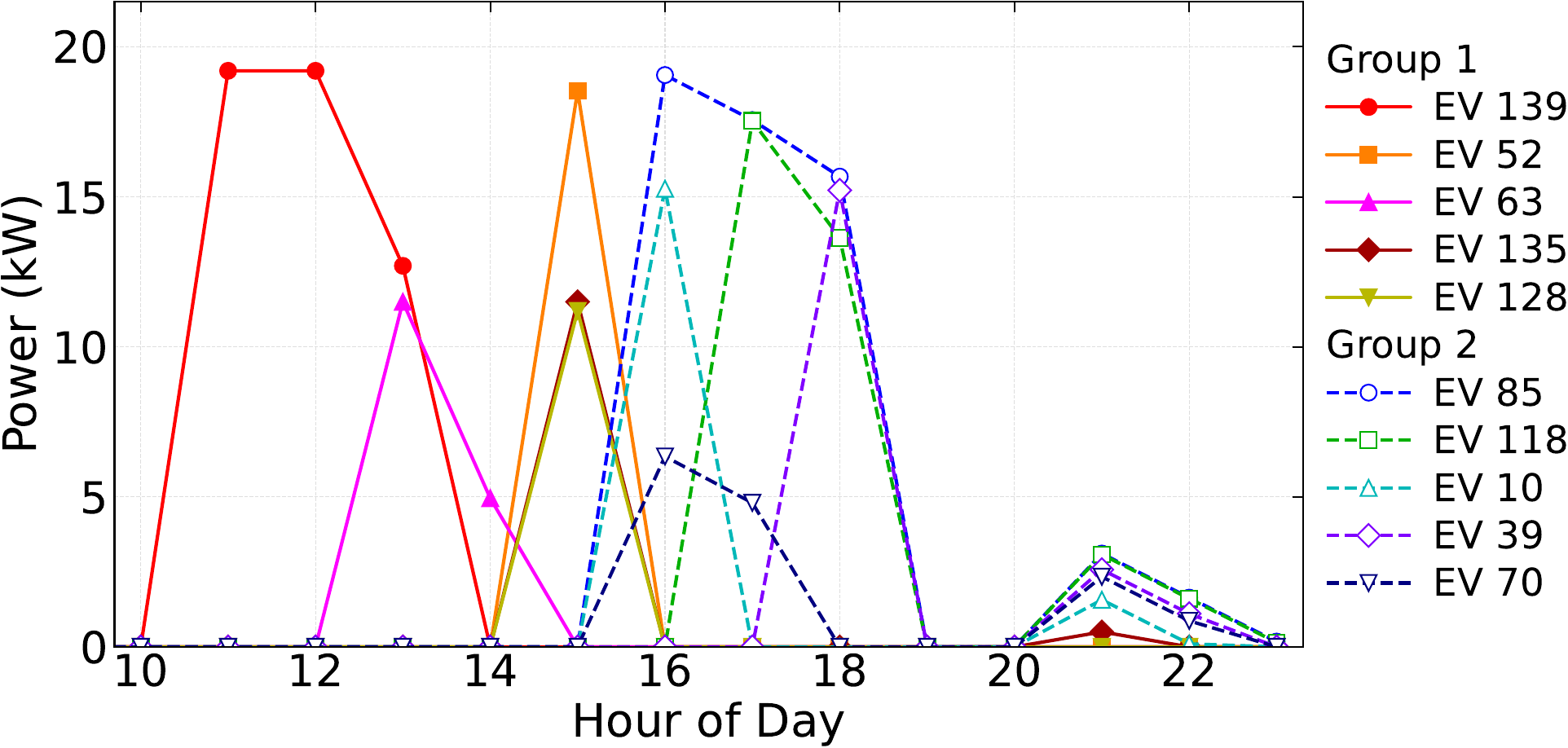}}\hspace{6pt}%
\subfloat[\label{fig:A1_byz}]{\includegraphics[height=\dimexpr(\linewidth-13pt)*1000/6190\relax]{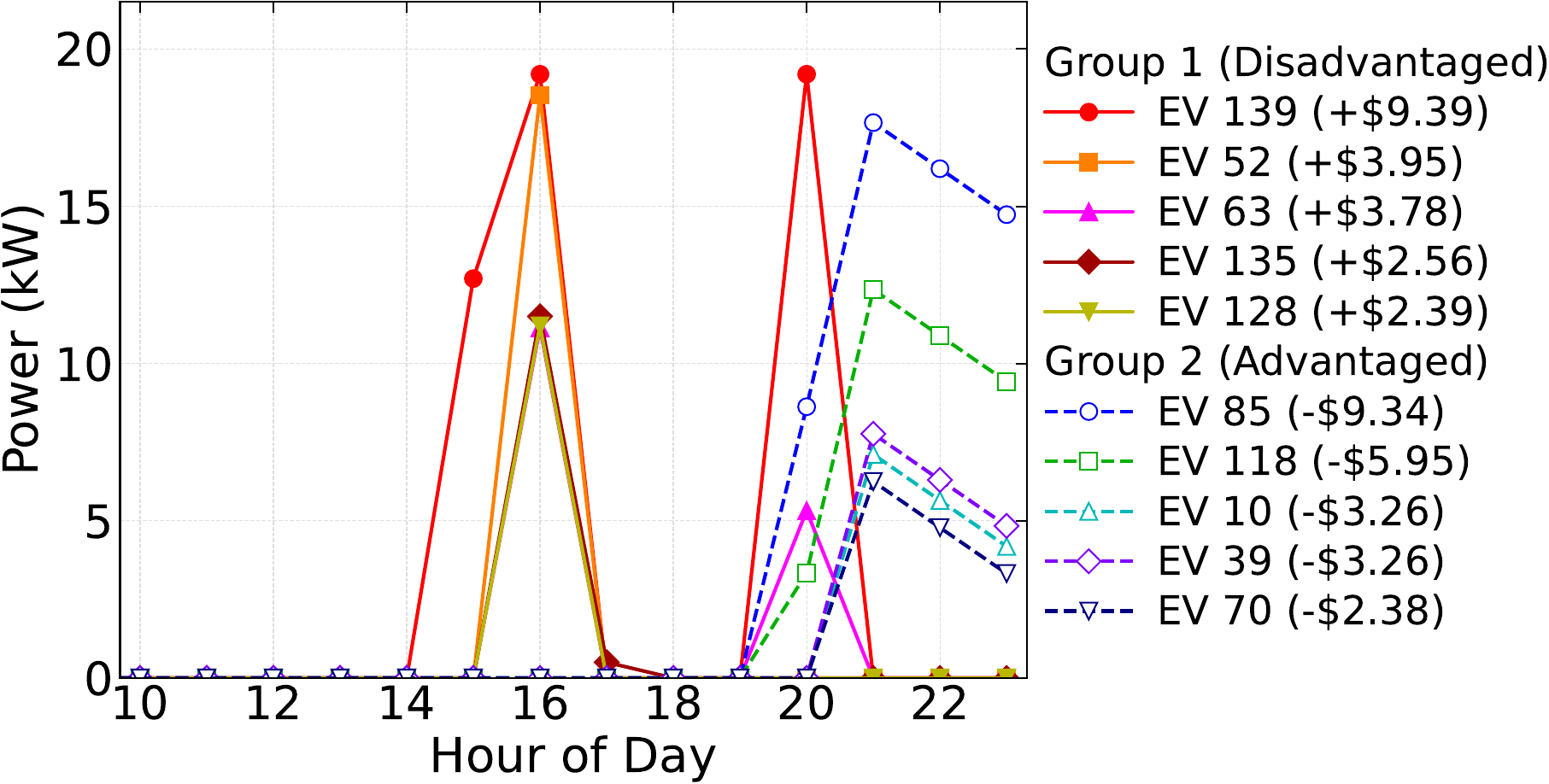}}\hspace{6pt}%
\subfloat[\label{fig:A1_rec}]{\includegraphics[height=\dimexpr(\linewidth-13pt)*1000/6190\relax]{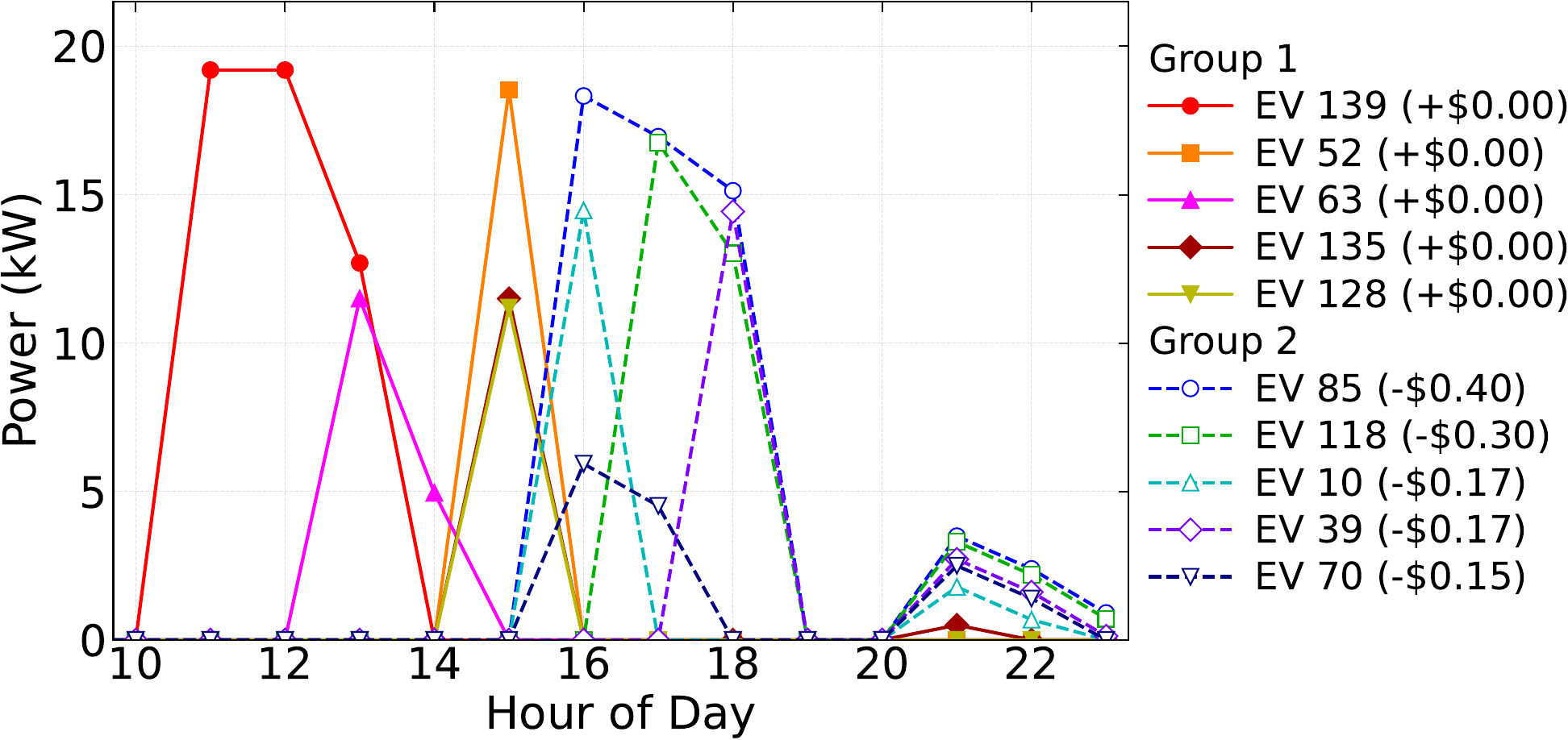}}\\[-2pt]
\subfloat[\label{fig:A2_ben}]{\includegraphics[height=\dimexpr(\linewidth-13pt)*1000/6191\relax]{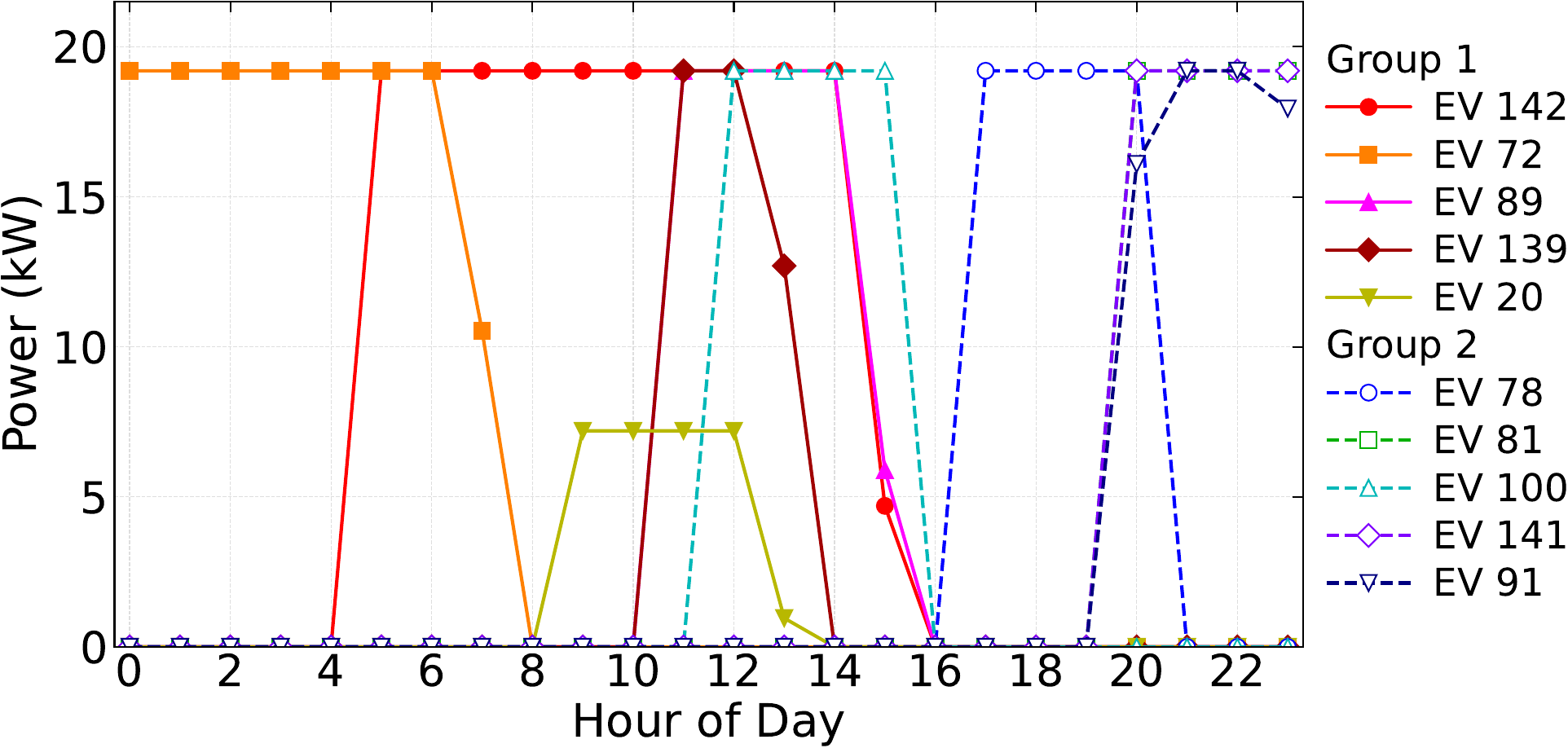}}\hspace{6pt}%
\subfloat[\label{fig:A2_byz}]{\includegraphics[height=\dimexpr(\linewidth-13pt)*1000/6191\relax]{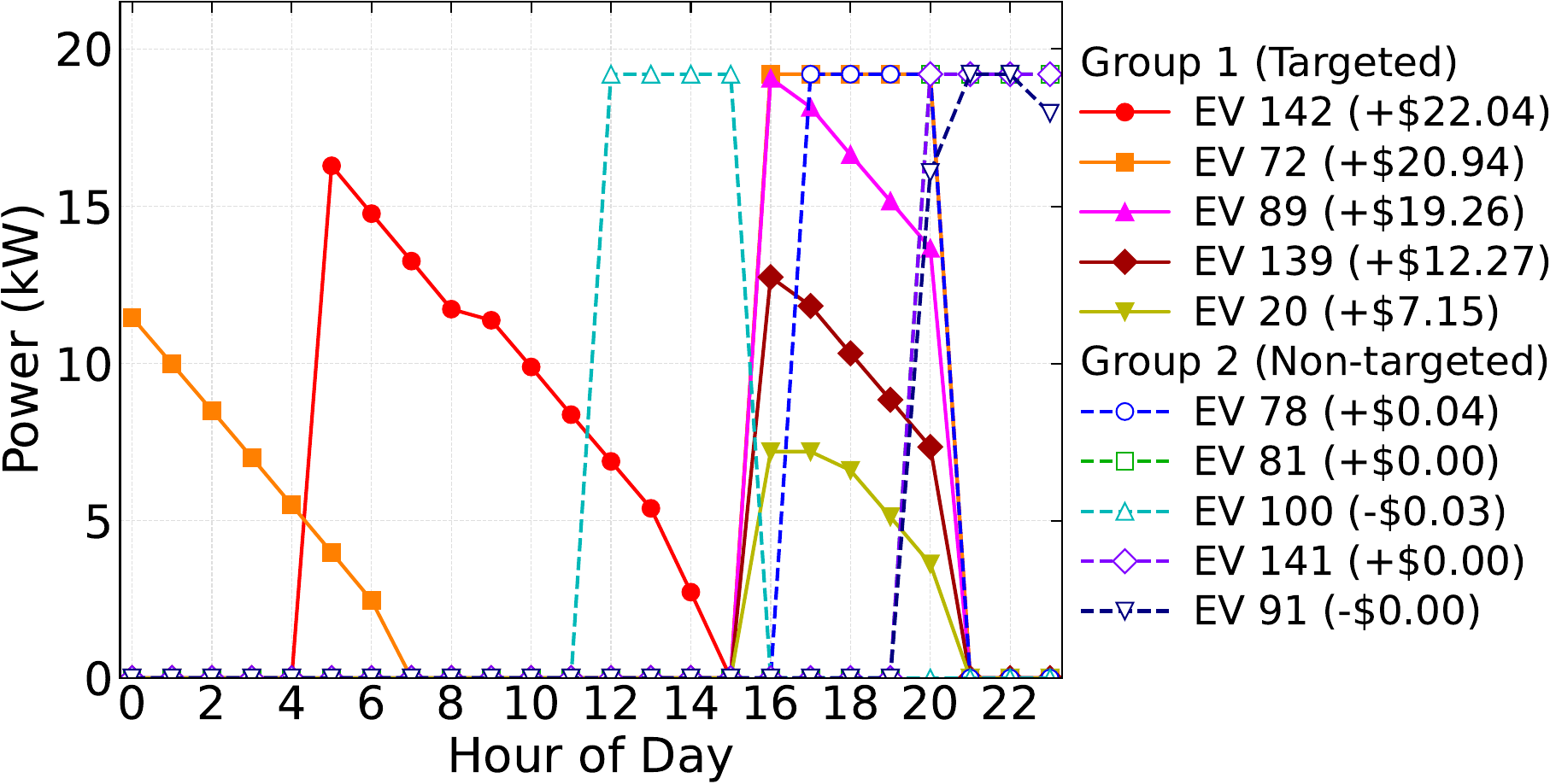}}\hspace{6pt}%
\subfloat[\label{fig:A2_rec}]{\includegraphics[height=\dimexpr(\linewidth-13pt)*1000/6191\relax]{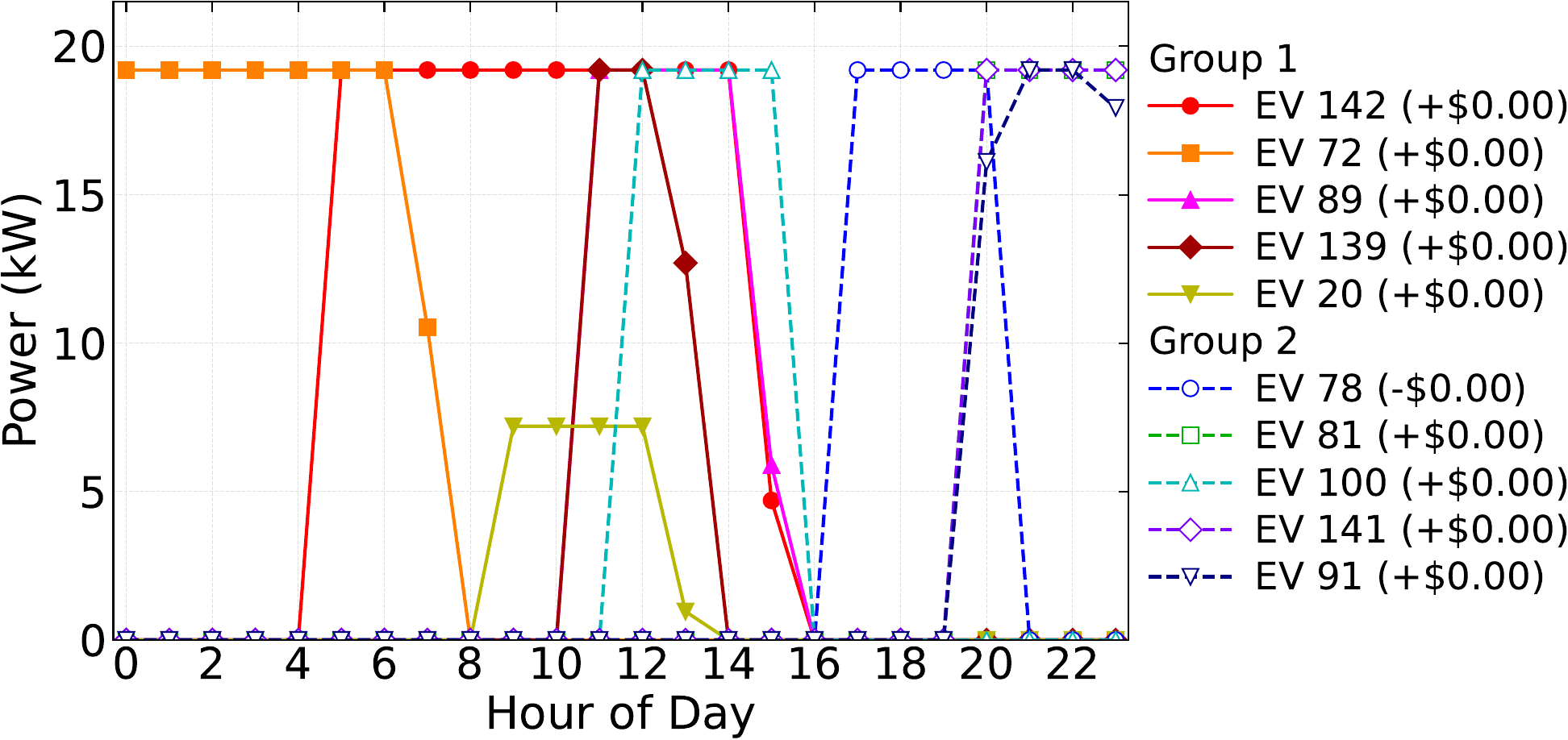}}
\caption{Charging schedules, with daily cost changes in parentheses: (a) benign, (b) attacked, and (c) recovered under A1; (d)--(f) the same under A2.}
\label{fig:schedules}
\end{figure*}

Fig.~\ref{fig:annual_cost_comparison} reports the annual cost per EV of each group. Under A1, the disadvantaged EVs pay 132.7\% more and the advantaged EVs 49.3\% less, whereas the non-attacked EVs are unaffected. Under A2, the targeted EVs pay 57.6\% more, and the cost of the non-targeted EVs changes by only 0.1\%. With the defense, all group costs remain within 2.6\% of benign operation. The largest residual, $-2.6\%$ for the advantaged EVs, arises because their recovered schedules follow the forecast rather than the final certified price. Fig.~\ref{fig:schedules} shows the benign, attacked, and recovered schedules of the five most disadvantaged or targeted EVs (Group~1) and of five EVs of the other group (Group~2). Without the defense, the Group~1 EVs are pushed into the peak-price hours and pay \$2.39 to \$22.04 more per day, whereas the advantaged EVs of A1 charge at their rated power in the cheapest hours and save up to \$9.34 per day. With the defense, every EV returns to a schedule close to its benign one, and no daily cost changes by more than \$0.40. The schedules also show realistic behavior, with each EV charging near its rated power for about $E_i/p_i^{\max}$ hours.

Table~\ref{tab:summary_results} summarizes the system metrics.
\begin{table}[!t]
\centering
\renewcommand{\arraystretch}{1.05}
\scriptsize
\setlength{\tabcolsep}{2.4pt}
\caption{Case-Study Results Without and With the Proposed Defense}
\label{tab:summary_results}
\begin{tabular}{lccccc}
\toprule
\multirow{2}{*}{\textbf{Metric}} & \multirow{2}{*}{\textbf{Benign}} & \multicolumn{2}{c}{\textbf{A1: Biased operator}} & \multicolumn{2}{c}{\textbf{A2: MITM attacker}} \\
\cmidrule(lr){3-4}\cmidrule(lr){5-6}
& & \textbf{Undefended} & \textbf{Defended} & \textbf{Undefended} & \textbf{Defended} \\
\midrule
Change of $J$ (\%)          & 0.0    & $-0.2$ & $-0.1$ & 15.8   & $-0.3$ \\
LFS (\%)                    & 3.9    & 5.6    & 4.0    & $-1.6$ & 4.1 \\
Peak load change (\%)       & 0.0    & $-1.7$ & $-0.1$ & 5.8    & $-0.1$ \\
Total EV payments (\$)      & 928.6  & 926.5  & 927.4  & 1113.3 & 925.7 \\
Station profit (\$)         & 43.9   & 43.9   & 43.9   & 49.8   & 43.9 \\
Iterations                  & 115    & 106    & 105    & 167    & 84 \\
Detection events            & 0      & 0      & 10     & 0      & 40 \\
Convergence time (s)        & 0.002  & 0.002  & 1.62   & 0.004  & 1.29 \\
Payload (B/EV/iteration)    & 384    & 384    & 722    & 384    & 722 \\
\bottomrule
\end{tabular}
\end{table}
Under A1, $J$ changes by $-0.2\%$ and the station profit by $-0.1\%$, which confirms that the attack is invisible in aggregate indicators. Under A2, $J$ increases by 15.8\% and the LFS turns negative, because the victims charge during the peak. The station profit increases by 13.4\%, because the victims buy energy in congested hours, in which the price exceeds the tariff by a congestion premium that the station retains. The defended runs keep $J$ within 0.3\% of its benign value. Since $J$ omits the preference costs of the EVs, the relevant measure is $J_{\mathrm{tot}}$, which exceeds its benign value by only \$0.08 per day under A1 and \$0.17 under A2, far below the corresponding bounds of Theorem~\ref{thm:cost_recovery}, \$4.91 and \$27.89. Each victim detected the attack in the onset iteration, which produced 10 and 40 detection events, and benign operation produced none, as Corollary~\ref{cor:detection} states. The two defended columns differ because the attacks target different EVs; for a given erasure pattern, the defended outcome is independent of the falsified content (Corollary~\ref{cor:content}).

\subsection{Comparison of Recovery Rules}
\label{subsec:recovery_results}
\begin{figure}[!t]
\centering
\subfloat[\label{fig:Convergence_rate}]{\includegraphics[width=0.49\linewidth]{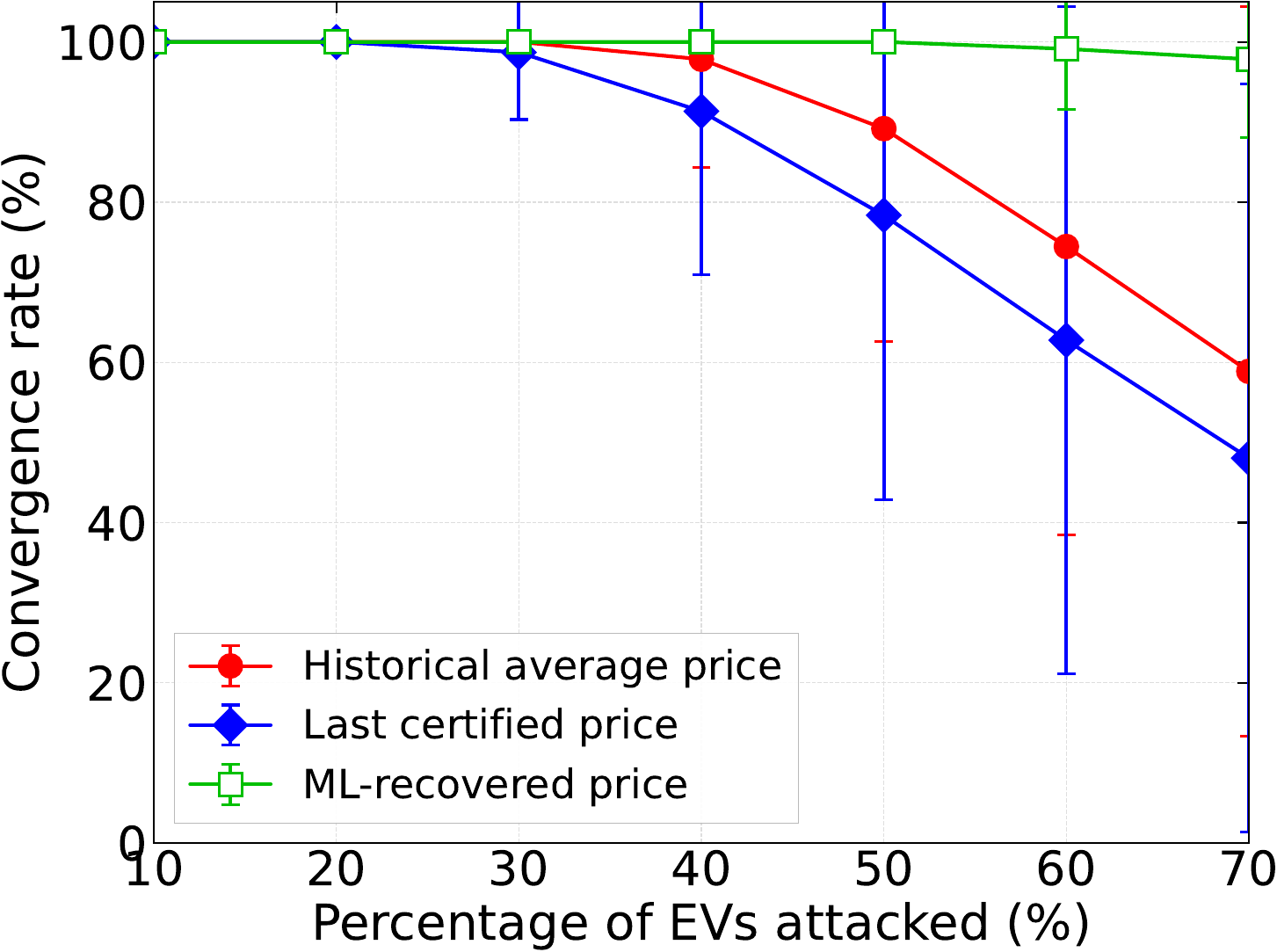}}\hspace{2pt}%
\subfloat[\label{fig:Att_cost}]{\includegraphics[width=0.49\linewidth]{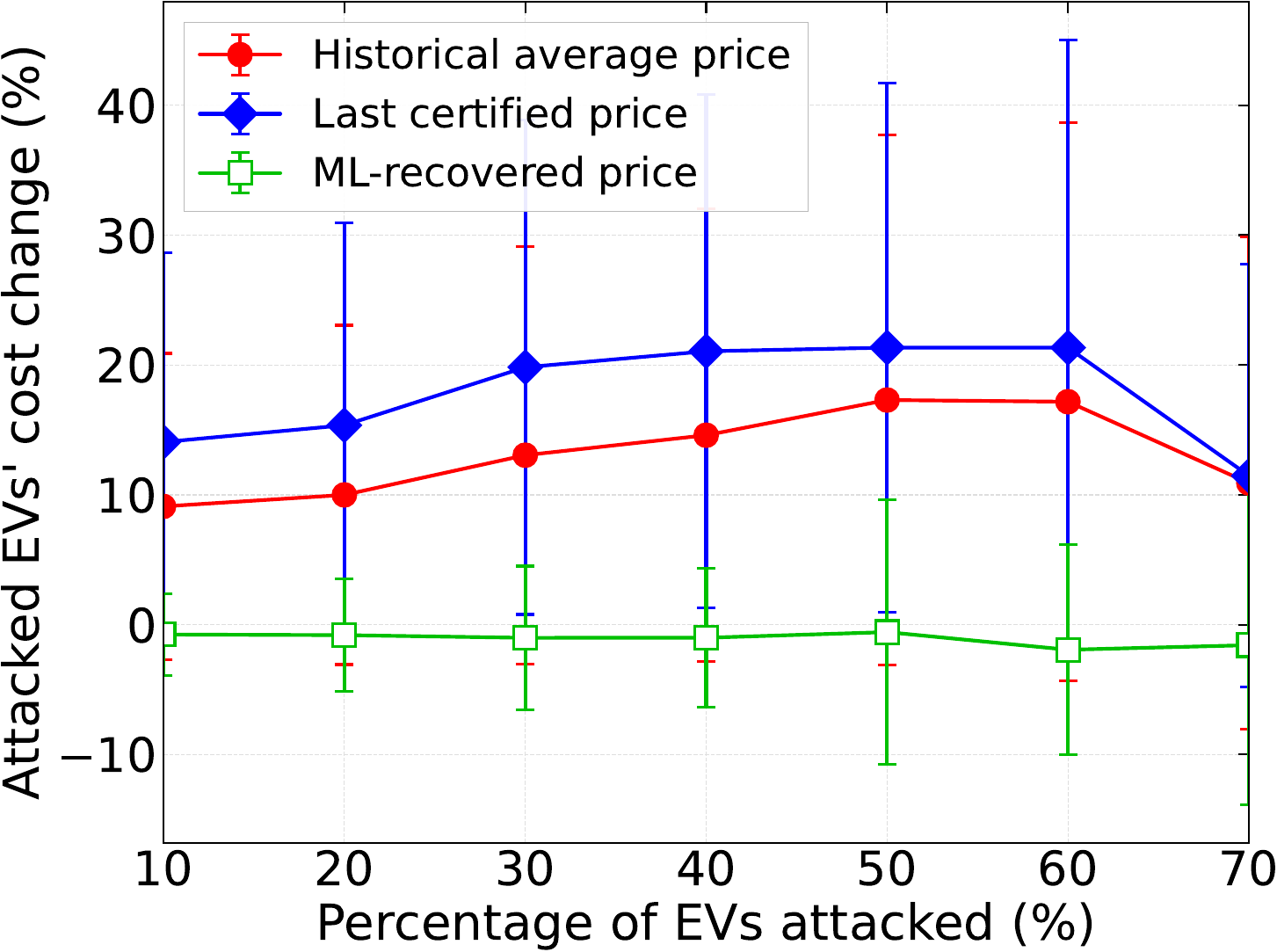}}
\caption{Comparison of recovery rules over 77 held-out days: (a) convergence rate and (b) cost change of the attacked EVs. Markers show means, and bars show standard deviations over the days.}
\label{fig:recovery_methods_comparison}
\vspace{5pt}
\end{figure}

To evaluate the recovery beyond a single day, the defense was applied on all 77 held-out days, whose fleets range from 25 to 163 EVs, with 10\% to 70\% of the EVs attacked. For each day and attack share, three attacked sets were drawn at random, and every attack started at iteration 2, long before convergence, which is the most demanding case for recovery. The attacked sets were drawn independently of the forecaster, which was trained offline and therefore had no advance knowledge of the victims. The proposed recovery was compared with the same rule \eqref{eq:recovery} driven by the historical average price or by the last certified price before the detection event. Fig.~\ref{fig:recovery_methods_comparison} reports the convergence rate and the change in the payments of the attacked EVs, as means and standard deviations over the days, and the text below gives the cost gap $J_{\mathrm{tot}}-J_{\mathrm{tot}}^\ast$ over the converged runs.
A run counts as converged when both residuals fall below their tolerances within 3,000 iterations, as in all other experiments. Each attack share thus comprises 231 runs, three attacked sets on each of the 77 days, and runs that do not converge count against the convergence rate but are excluded from the cost-gap statistics reported below.

The proposed recovery converges in all runs up to an attack share of 50\% and in 99.1\% and 97.8\% of the runs at 60\% and 70\%, respectively.
\begin{figure*}[!b]
\centering
\subfloat[\label{fig:scal_iter}]{\includegraphics[height=\dimexpr(\linewidth-13pt)*1000/4092\relax]{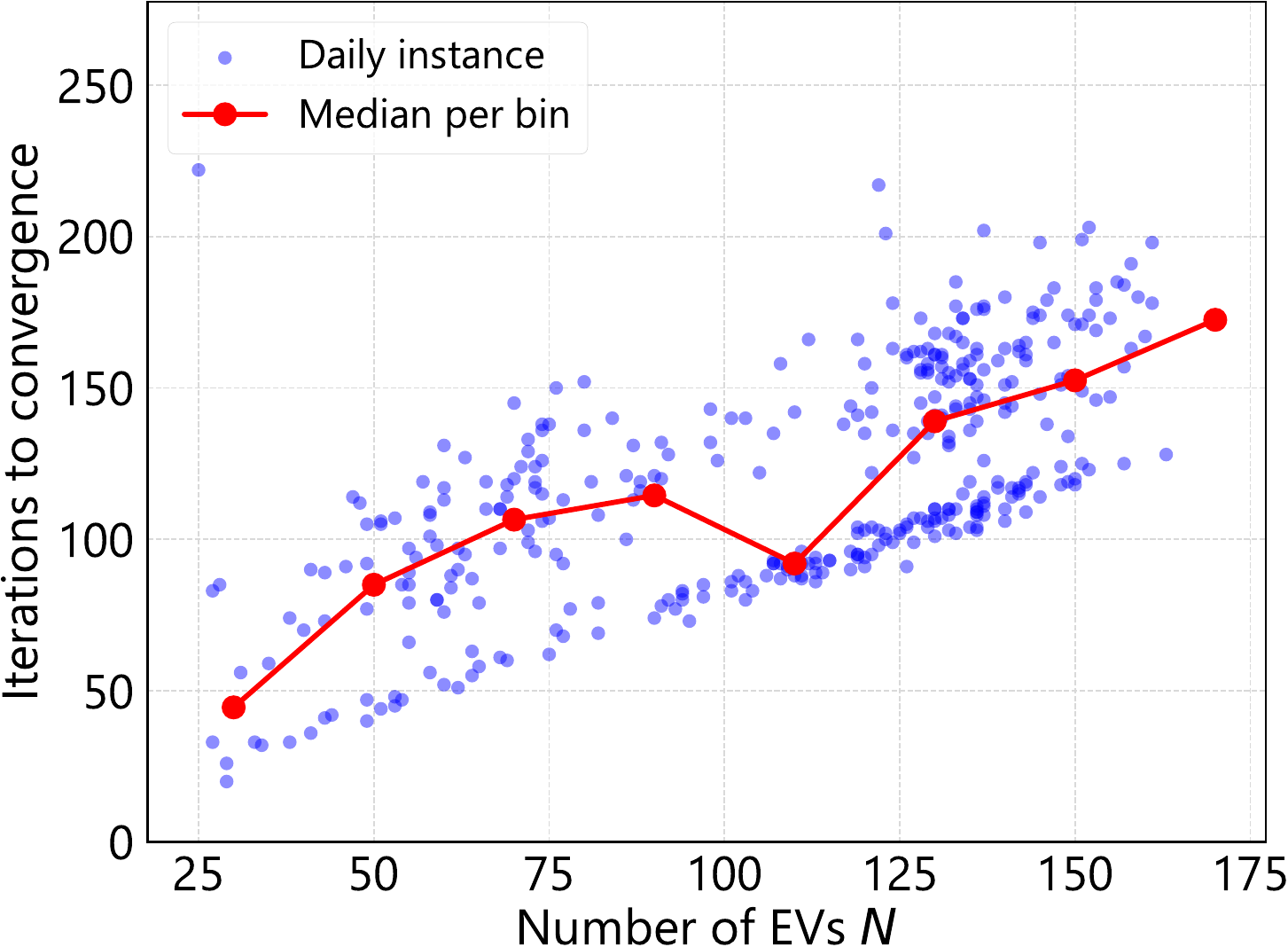}}\hspace{6pt}%
\subfloat[\label{fig:scal_lat}]{\includegraphics[height=\dimexpr(\linewidth-13pt)*1000/4092\relax]{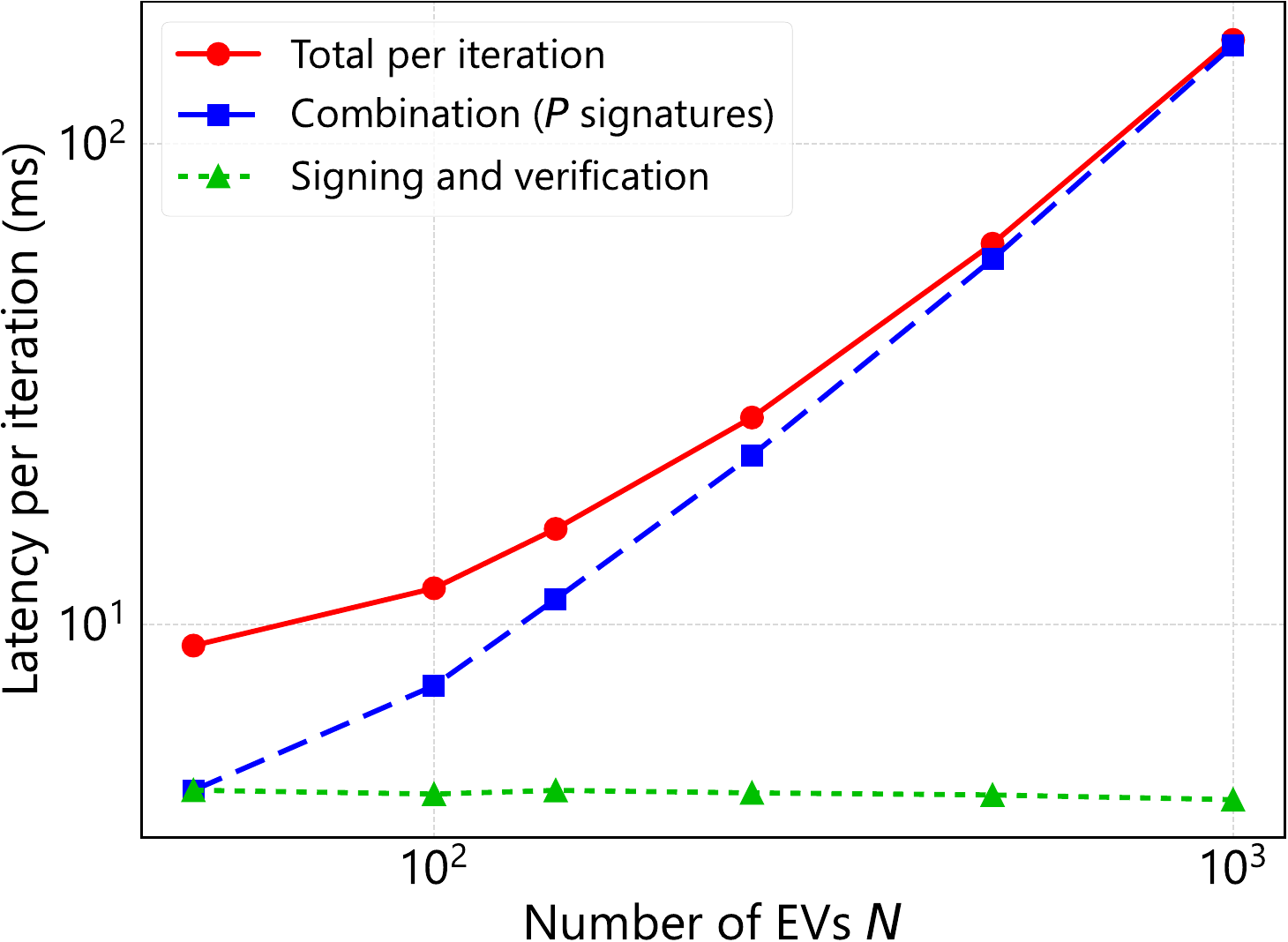}}\hspace{6pt}%
\subfloat[\label{fig:scal_msg}]{\includegraphics[height=\dimexpr(\linewidth-13pt)*1000/4092\relax]{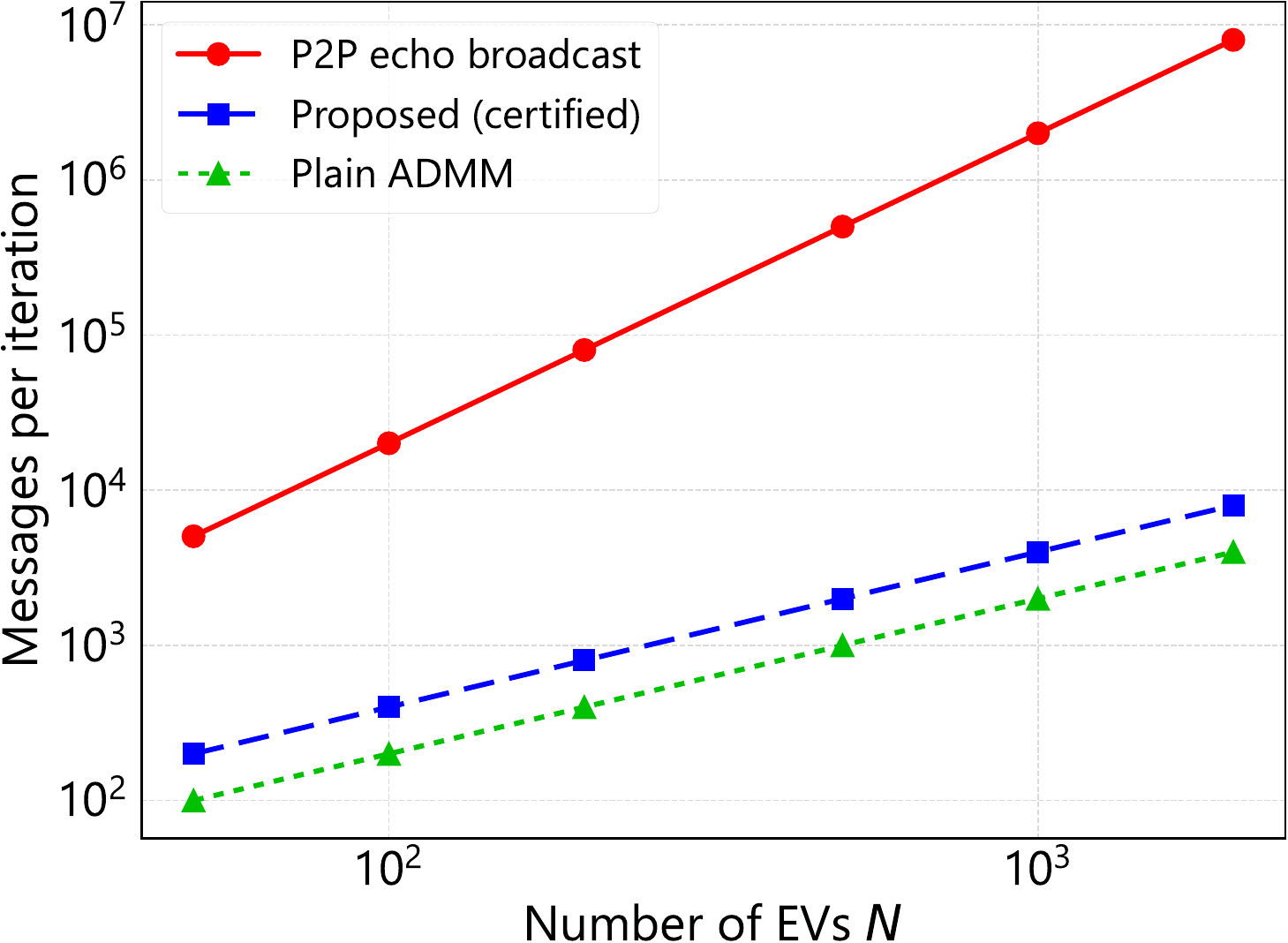}}
\caption{Scalability of the defense: (a) iterations to convergence over all 365 daily instances, (b) latency of a certified iteration, and (c) messages per iteration.}
\label{fig:scalability}
\end{figure*}
With the historical average, the convergence rate falls to 89.2\% at 50\% and 58.9\% at 70\%, and with the last certified price, it falls to 78.4\% and 48.1\%. Theorem~\ref{thm:conv_recovery} explains these failures. Both baselines miss the congestion hours of the day. The recovered EVs therefore place their fixed load in slots without headroom, \eqref{eq:feasibility} is violated, and $\mathcal{P}_\Aset$ becomes infeasible, which the operator observes as a primal residual that does not vanish. The last certified price converges least often, because at an early onset the iterate is still far from the benign price. The fleet-aware forecast anticipates congestion and therefore keeps the recovered schedules feasible.

\begin{table}[!t]
\centering
\renewcommand{\arraystretch}{1.05}
\scriptsize
\setlength{\tabcolsep}{2.6pt}
\caption{Computation and Communication Cost per Iteration ($N=142$)}
\label{tab:cost}
\begin{tabular}{lccc}
\toprule
\textbf{Quantity} & \textbf{Plain ADMM} & \textbf{Proposed} & \textbf{P2P echo} \\
\midrule
Messages & $2N=284$ & $4N=568$ & $2N^2+N=40{,}470$ \\
Payload per EV (B) & 384 & 722 & $384(2N+1)$ \\
Data volume (kB) & 54.5 & 102.5 & 15,540 \\
EV computation & 4.3~$\mu$s & 2.5~ms & $N$ checks \\
Operator computation & 17.0~$\mu$s & 12.8~ms & -- \\
Latency per iteration & 0.021~ms & 15.4~ms & -- \\
Links among EVs & None & None & $N(N-1)$ \\
\bottomrule
\end{tabular}
\vspace{1.5pt}
\end{table}

For attack shares of 10\% to 50\%, the cost gap of the proposed recovery ranges from \$0.51 to \$3.69 per day, compared with \$3.88 to \$26.83 for the historical average and \$5.73 to \$30.44 for the last certified price, i.e., 7.3 to 8.1 times lower than with the better baseline. The proposed recovery also keeps the payments of the attacked EVs within 2\% of their benign values, whereas the historical average raises them by 9.1\% to 17.3\% and the last certified price by 14.1\% to 21.3\%. In agreement with Theorem~\ref{thm:cost_recovery}, its bound held in all converged runs of the proposed recovery, with a median ratio of 0.023 between the measured gap and the bound. The sufficient condition \eqref{eq:feasibility} held in only 40\% to 45\% of these runs and is therefore conservative.

\subsection{Overhead and Scalability}
\label{subsec:overhead_results}
Table~\ref{tab:cost} compares the computation and communication cost per iteration of plain ADMM, of the proposed certified iteration, and of P2P echo broadcast~\cite{Bracha1987Reliable}, the classical means of detecting inconsistent broadcasts without trusting the coordinator, and Fig.~\ref{fig:scalability} evaluates the scalability of the defense. The latency assumes that the EVs compute, sign, and verify in parallel, and each of these operations therefore contributes once per iteration. In the case study, an EV update takes 4.3~$\mu$s, a partial signature 0.68~ms, and a verification 1.86~ms, whereas the operator combines the $P=95$ partial signatures in 11.0~ms. A certified iteration therefore takes 15.4~ms, and the defended runs converge within 1.7~s, which is negligible for day-ahead scheduling. These latencies exclude network delays; compared with plain ADMM, certification adds one round trip between the operator and the signers to each iteration. In benign operation, the certified iteration needs the same 115 iterations as plain ADMM, because certification does not change the updates, and the ML forecaster, which is trained and certified offline, runs only after a detection event. Certification nearly doubles the payload per EV, but the number of messages remains linear in $N$, whereas echo broadcast requires $2N^2+N$ messages, i.e., 40,470 instead of 568 for $N=142$, and it needs links among all EVs, unlike the proposed iteration.

Over the 365 daily instances, the median number of iterations grows moderately, from about 45 for 30 EVs to about 170 for 170 EVs, and its scatter reflects the different energy requirements and windows of the daily fleets. The latency of a certified iteration grows linearly with $N$, from 9.0~ms for 50 EVs to 164.7~ms for 1,000 EVs, because the operator combines $P\approx2N/3$ partial signatures, whereas signing and verification take constant time. A session with 1,000 EVs and a few hundred iterations therefore completes within one minute. Meanwhile, the number of messages remains $4N$, three orders of magnitude below that of echo broadcast for 1,000 EVs. A committee of $M$ institutions, with $2N+2M$ messages per iteration, makes the combination cost independent of $N$ for larger fleets, because the operator then combines only the $P$ partial signatures of the $M$ members.

\section{Discussion and Limitations}
\label{sec:limitations}
The forecaster generalizes across fleet sizes because its inputs describe each day's fleet relative to the station rating, and its test days span 25 to 163 EVs. A station with a different tariff, rating, or EV population can retrain it on its own certified sessions, which accumulate during benign operation. If the inputs drift beyond the training range, the error span grows, which by Theorem~\ref{thm:cost_recovery} increases the cost bound without affecting detection or convergence, and the drift shows in the forecast errors of recent sessions, which allows the signers to refuse recertifying an outdated model.

The protection also extends to other coordinator-based iterations, such as dual decomposition or consensus ADMM, and to other flexible loads, such as heat pumps and stationary batteries, because certification wraps only the broadcast step and recomputation requires only deterministic updates of certified data. It also preserves privacy, as it only adds signatures to plain ADMM messages and reveals no further EV data. An undetected attack forfeits the benefit of coordination, because a MITM attacker pushes its victims into the peak-price hours that coordination avoids, and a biased operator redistributes the costs that coordination allocates. Coordination is also needed on light days, because on the tenth of the days with the least energy, 41.2\% of the uncoordinated draws exceed the station rating, as the sessions of fewer EVs average out less. Field deployment would also require an EA supported by EV public-key infrastructures, with manufacturers or charging providers operating EV-side agents, so that drivers need no additional equipment and interact with the station as before.

The study has four limitations. First, no public dataset jointly records charging sessions, bus load, prices, and station rating. The data therefore originate from different sources, and the experiments constitute a California-based case study rather than a reconstruction of a specific site. Second, public datasets with the arrival and departure times, energy requirements, and charging power of individual vehicles are scarce and small. The largest daily fleet available to this study contained 163 EVs, and larger fleets were assessed through measurements of computation and communication cost. Third, annual costs are indicative, based on one day’s price per kWh and typical annual charging energy, as EVs charge intermittently at the observed site. Fourth, the study optimizes a single charging station. Multiple stations can coordinate their EVs with the same method, and future work will develop a secure collaboration layer among stations to reduce their combined grid impact. 

\section{Conclusion}
\label{sec:conclusion}
This paper showed that inconsistency attacks by an operator or a MITM attacker, together with EVs that deviate from their updates, converge to a perturbed optimum without any observable anomaly, and neither residual monitoring nor the published price reveals them. Embedding a threshold-certified broadcast, whose key is generated through the untrusted operator and anchored only by an offline enrollment of signers, into every ADMM iteration and recomputing every upload turns each such attack into a deterministic detection event, without P2P links and with linear communication. Because detection relies only on standard cryptographic assumptions, it does not trade missed detections against false alarms.

Detection alone, however, would stall the optimization. A certified, fleet-aware price forecast lets each affected EV recover locally, keeps the optimization convergent and feasible from any attack onset, and bounds the extra cost by the shape of the forecast error rather than its level. The simulations first confirmed that coordination itself is essential, because without it EV charging accumulates in the peak-price hours, raises the charging cost on every day, and can overload the station. This benefit grows with the charging demand, which makes a secure coordination layer increasingly important as EV adoption rises. They further showed that a biased coordinator can redistribute costs among EVs without any aggregate trace, that an external attacker can additionally stress the grid in peak hours, and that the defense restores near-benign operation in both cases, with a recovery that remains reliable even when most EVs are attacked. Certification also keeps the communication linear in the fleet size, and a session with 1,000 EVs completes within one minute. Because the certification leaves the optimization model unchanged, it can be added to existing ADMM-based stations without redesigning their schedules. Future work will develop a secure collaboration layer among stations and extend the framework to V2G schedules.

\appendices
\section{Proofs}
\label{app:proofs}
We first state an inequality used in several proofs. Every $\mathbf{x}\in\mathcal{X}_i$ vanishes outside $\Tset_i$ and satisfies $\mathbf{1}^\top\mathbf{x}=E_i/\Delta t$. Hence, for any $\mathbf{x},\mathbf{x}'\in\mathcal{X}_i$, any vector $\mathbf{w}$, and any scalar $\vartheta$, $\mathbf{w}^\top(\mathbf{x}-\mathbf{x}')=\sum_{t\in\Tset_i}(w(t)-\vartheta)(x(t)-x'(t))$, and choosing $\vartheta$ as the midpoint of the range of $\mathbf{w}$ on $\Tset_i$ and using $\|\mathbf{x}-\mathbf{x}'\|_1\le2E_i/\Delta t$ gives
\begin{equation}
\mathbf{w}^\top(\mathbf{x}-\mathbf{x}')\le\tfrac{1}{2}\mathrm{sp}_{\Tset_i}(\mathbf{w})\,\|\mathbf{x}-\mathbf{x}'\|_1\le\frac{E_i}{\Delta t}\,\mathrm{sp}_{\Tset_i}(\mathbf{w}).
\label{eq:span_lemma}
\end{equation}
The same identity shows that adding $\vartheta\mathbf{1}$ to $\mathbf{w}$ changes $\mathbf{w}^\top\mathbf{x}$ by the constant $\vartheta E_i/\Delta t$ on $\mathcal{X}_i$.

\noindent\emph{Proof of Lemma~\ref{lem:benign}:} Problem \eqref{eq:obj_site}--\eqref{eq:site_rating} reads $\min F(\mathbf{x})+G(\mathbf{x})$ with $F(\mathbf{x})=\sum_ih_i(\mathbf{x}_i)$ and $G(\mathbf{y})=(J+\iota_{\mathcal{Z}})(\sum_i\mathbf{y}_i)$, both closed proper convex, and has a KKT point under Assumption~\ref{ass:feasible}; iteration \eqref{eq:v_def_unscaled}--\eqref{eq:lambda_update_site} is ADMM for it with $\mathbf{S}=N\bar{\mathbf{x}}$, $\mathbf{z}=N\bar{\mathbf{y}}$, and $\boldsymbol{\lambda}=\rho\mathbf{u}$~\cite[Sec.~3.3]{Boyd2011ADMM}, and it therefore converges~\cite[Thm.~8]{Eckstein1992DR}. For (a), $\mathbf{S}^\ast=\mathbf{z}^\ast$ gives $\mathbf{v}^\ast=-\boldsymbol{\lambda}^\ast/\rho$, and $\mathbf{x}_i^\ast$ therefore minimizes $h_i(\mathbf{x})+\frac{\rho}{2}\|\mathbf{x}-\mathbf{x}_i^\ast+\boldsymbol{\lambda}^\ast/\rho\|_2^2$; its optimality condition $\mathbf{0}\in\partial h_i(\mathbf{x}_i^\ast)+\boldsymbol{\lambda}^\ast$ is that of minimizing $\boldsymbol{\lambda}^{\ast\top}\mathbf{x}+\varphi_i(\mathbf{x})$ over $\mathcal{X}_i$. For (b), the $\mathbf{z}$-update gives $\boldsymbol{\lambda}^\ast\in\nabla J(\mathbf{z}^\ast)+N_{\mathcal{Z}}(\mathbf{z}^\ast)$, and the normal cone $N_{\mathcal{Z}}$ vanishes in slots with $0<z^\ast(t)<P_{\mathrm{CS}}^{\max}$.\hfill$\blacksquare$

\noindent\emph{Proof of Proposition~\ref{prop:stealth}:} For $i\in\Aset$ and $k\ge k_0$, \eqref{eq:ev_update_v_only_attacked} gives $\mathbf{x}_i^{k+1}=\arg\min_{\mathbf{x}\in\mathcal{X}_i}\{\varphi_i(\mathbf{x})-\rho\boldsymbol{\delta}_i^\top\mathbf{x}+\frac{\rho}{2}\|\mathbf{x}-\mathbf{x}_i^k-\mathbf{v}^k\|_2^2\}$, which is the update \eqref{eq:sharing_x} with $h_i(\mathbf{x})=\iota_{\mathcal{X}_i}(\mathbf{x})+\varphi_i(\mathbf{x})-\rho\boldsymbol{\delta}_i^\top\mathbf{x}$, and all other updates are unchanged. Hence, from $k_0$ on, the iteration is sharing-form ADMM for the feasible convex QP \eqref{eq:perturbed_problem}, and the argument of Lemma~\ref{lem:benign} gives (a) and (c). For (b), $\tilde{\mathbf{S}}=\tilde{\mathbf{z}}$ at the limit, and the argument of Lemma~\ref{lem:benign}(a) shows that $\tilde{\mathbf{x}}_i$ minimizes $(\tilde{\boldsymbol{\lambda}}-\rho\boldsymbol{\delta}_i)^\top\mathbf{x}+\varphi_i(\mathbf{x})$ over $\mathcal{X}_i$ for $i\in\Aset$.\hfill$\blacksquare$

\noindent\emph{Proof of Corollary~\ref{cor:damage}:} By Proposition~\ref{prop:stealth}(b), $(\tilde{\boldsymbol{\lambda}}-\rho\boldsymbol{\delta}_i)^\top\tilde{\mathbf{x}}_i+\varphi_i(\tilde{\mathbf{x}}_i)\le(\tilde{\boldsymbol{\lambda}}-\rho\boldsymbol{\delta}_i)^\top\bar{\mathbf{x}}_i+\varphi_i(\bar{\mathbf{x}}_i)$; hence, $\tilde C_i(\tilde{\mathbf{x}}_i)-\tilde C_i(\bar{\mathbf{x}}_i)\le\rho\boldsymbol{\delta}_i^\top(\tilde{\mathbf{x}}_i-\bar{\mathbf{x}}_i)$. Since $|\gamma_i|=1$, $\mathrm{sp}_{\Tset_i}(\rho\boldsymbol{\delta}_i)=\rho\tau_i\mathrm{sp}_{\Tset_i}(\mathbf{a}_i)$, and \eqref{eq:span_lemma} with $\mathbf{w}=\rho\boldsymbol{\delta}_i$ gives \eqref{eq:damage}; \eqref{eq:span_lemma} with $\mathbf{w}=\tilde{\boldsymbol{\lambda}}$ bounds the extra payment. If $a_i(t)=+1$ and $a_i(s)=-1$ for $t,s\in\Tset_i$, then $(\tilde\lambda(t)-\rho\tau_i)-(\tilde\lambda(s)+\rho\tau_i)\le\mathrm{sp}_{\Tset_i}(\tilde{\boldsymbol{\lambda}})-2\rho\tau_i<0$, and a linear price is minimized by filling the cheapest slots first.\hfill$\blacksquare$

\noindent\emph{Proof of Lemma~\ref{lem:dkg}:} (a) Suppose that two transcripts with digests $h\neq h'$ are certified in one phase. Barring a forged credential signature or a digest collision, each carries signatures of at least $P$ roster members, and an honest member signs at most one digest per phase; hence, the honest signers of $h$ and $h'$ are disjoint sets of at least $P-f$ members each, which requires $2(P-f)\le N_s-f$, i.e., $P\le(N_s+f)/2$, a contradiction. Since $\mathcal{Q}$, $\mathrm{pk}$, and the public shares are functions of the certified transcripts, all honest signers agree on them. (b) The certified sharing-phase transcript carries at least $P-f\ge1$ honest signatures, and an honest member signs only if $|\mathcal{Q}|\ge f+1$; hence, $\mathcal{Q}$ contains an honest dealer; honest dealers publish their extraction values only after $\mathcal{Q}$ is certified, as in~\cite{Gennaro2007DKG}, and the adversary therefore cannot choose $\mathcal{Q}$ after seeing them. (c) If the operator forwards all messages, every responding honest member receives a transcript that contains its messages, at least $N_s-f-d\ge P>f$ honest dealers qualify, and at least $N_s-f-d\ge P$ signatures arrive.\hfill$\blacksquare$

\noindent\emph{Proof of Theorem~\ref{thm:byz_collusion}:} Consider the executions without a hash collision; adding $\epsilon_H$ completes the proof. (i) If an honest EV accepts a certified message that fewer than $P-|\mathcal{C}|$ honest signers signed, the adversary, holding $|\mathcal{C}|<P$ shares, has produced a valid threshold signature without the required honest partial signatures, i.e., a threshold BLS forgery, which occurs with probability at most $\epsilon_{\mathrm{BLS}}$. If two honest EVs accept $m\neq m'$ for the same $(\Sset,k)$ without such a forgery, each message received partial signatures from at least $P-|\mathcal{C}|$ honest signers, and since an honest signer signs at most one candidate per $(\Sset,k)$, these sets are disjoint, which implies $P\le(N_s+|\mathcal{C}|)/2\le(N_s+f)/2$, a contradiction. Hence, one certificate is a forgery, which again occurs with probability at most $\epsilon_{\mathrm{BLS}}$; by Lemma~\ref{lem:dkg}(a), all honest EVs verify under the same $\mathrm{pk}$. (ii) If the operator follows the protocol, at least $N_s-f-d\ge P$ available honest signers sign the unique candidate. (iii) The adversary holds $|\mathcal{C}|\le f<P$ shares, and by Lemma~\ref{lem:dkg}(b), $\mathcal{Q}$ contains an honest dealer whose secret stays hidden until $\mathcal{Q}$ is fixed. As in~\cite{Gennaro2007DKG}, the adversary's view can then be simulated from $\mathrm{pk}=\mathrm{sk}\,G$ alone, up to a negligible error, and outputting $\mathrm{sk}$ amounts to computing the discrete logarithm of $\mathrm{pk}$, possible with probability at most $\epsilon_{\mathrm{DL}}$.\hfill$\blacksquare$

\noindent\emph{Proof of Theorem~\ref{thm:conv_recovery}:} For $k>\bar k$, every $i\in\Aset$ uploads $\hat{\mathbf{x}}_i$, i.e., the update \eqref{eq:sharing_x} for $h_i=\iota_{\{\hat{\mathbf{x}}_i\}}$, and every other EV accepts the certified broadcast, which is the same for all honest EVs by Theorem~\ref{thm:byz_collusion}; hence, from $\bar k+1$ on, the iteration is sharing-form ADMM for the convex QP $\mathcal{P}_\Aset$, and if it is feasible, the argument of Lemma~\ref{lem:benign} gives convergence. If \eqref{eq:feasibility} holds, the schedules $\mathbf{x}_j^\ast$ ($j\notin\Aset$) and $\hat{\mathbf{x}}_i$ ($i\in\Aset$) lie in their sets, with a nonnegative aggregate of at most $P_{\mathrm{CS}}^{\max}$; hence, $\mathcal{P}_\Aset$ is feasible. Both sufficient conditions imply \eqref{eq:feasibility}, since $\hat x_i(t)-x_i^\ast(t)\le p_i^{\max}-x_i^\ast(t)$ and since $\sum_{i=1}^Np_i^{\max}\le P_{\mathrm{CS}}^{\max}$ gives $P_{\mathrm{CS}}^{\max}-z^\ast(t)\ge\sum_{i\in\Aset}(p_i^{\max}-x_i^\ast(t))$. Finally, if $\mathcal{P}_\Aset$ is infeasible but $r_\mathrm{p}^k\to0$ along a subsequence, compactness of $\mathcal{X}_i$ and $\mathcal{Z}$ yields a limit point with $\mathbf{S}=\mathbf{z}\in\mathcal{Z}$, $\mathbf{x}_i\in\mathcal{X}_i$, and $\mathbf{x}_i=\hat{\mathbf{x}}_i$ for $i\in\Aset$, which is feasible for $\mathcal{P}_\Aset$, a contradiction.\hfill$\blacksquare$

\noindent\emph{Proof of Theorem~\ref{thm:cost_recovery}:} \eqref{eq:span_lemma} with $\mathbf{w}=\boldsymbol{\lambda}^\ast$ gives \eqref{eq:envelope}, and the last statement of the span inequality shows that adding $\vartheta\mathbf{1}$ to $\hat{\boldsymbol{\lambda}}$ shifts the objective of \eqref{eq:recovery} by a constant on $\mathcal{X}_i$. Since $\hat{\mathbf{x}}_i$ minimizes \eqref{eq:recovery}, $\hat{\boldsymbol{\lambda}}^\top\hat{\mathbf{x}}_i+\varphi_i(\hat{\mathbf{x}}_i)\le\hat{\boldsymbol{\lambda}}^\top\mathbf{x}_i^\ast+\varphi_i(\mathbf{x}_i^\ast)+\frac{\tilde\rho}{2}\|\mathbf{x}_i^\ast-\mathbf{r}_i\|_2^2$, and adding $(\boldsymbol{\lambda}^\ast-\hat{\boldsymbol{\lambda}})^\top(\hat{\mathbf{x}}_i-\mathbf{x}_i^\ast)$, which \eqref{eq:span_lemma} bounds by $E_i\epsilon_i/\Delta t$, proves \eqref{eq:cost_bound_ev}. For \eqref{eq:cost_bound_sys}, let $\boldsymbol{\Delta}=\sum_{i\in\Aset}(\hat{\mathbf{x}}_i-\mathbf{x}_i^\ast)$; by \eqref{eq:feasibility}, the schedules in the proof of Theorem~\ref{thm:conv_recovery}, with aggregate $\mathbf{z}^\ast+\boldsymbol{\Delta}$, are feasible for $\mathcal{P}_\Aset$; hence, $J_\Aset\le J(\mathbf{z}^\ast+\boldsymbol{\Delta})+\sum_{i\in\Aset}\varphi_i(\hat{\mathbf{x}}_i)+\sum_{j\notin\Aset}\varphi_j(\mathbf{x}_j^\ast)=J_{\mathrm{tot}}^\ast+\nabla J(\mathbf{z}^\ast)^\top\boldsymbol{\Delta}+\Delta t\sum_tq(t)\Delta(t)^2+\sum_{i\in\Aset}[\varphi_i(\hat{\mathbf{x}}_i)-\varphi_i(\mathbf{x}_i^\ast)]$. Since $0<z^\ast(t)<P_{\mathrm{CS}}^{\max}$, Lemma~\ref{lem:benign}(b) gives $\nabla J(\mathbf{z}^\ast)=\boldsymbol{\lambda}^\ast$, and the first and last terms sum to $\sum_{i\in\Aset}[C_i^\ast(\hat{\mathbf{x}}_i)-C_i^\ast(\mathbf{x}_i^\ast)]$, and \eqref{eq:cost_bound_ev}, or \eqref{eq:envelope} plus the preference difference for a frozen EV, completes the proof.\hfill$\blacksquare$

\noindent\emph{Proof of Theorem~\ref{thm:baselines}:} (a) The objective of \eqref{eq:recovery} is $\tilde\rho$-strongly convex, and comparing its value at $\mathbf{r}_i\in\mathcal{X}_i$ with that at its minimizer gives $\hat{\boldsymbol{\lambda}}^\top\boldsymbol{\xi}+\varphi_i(\hat{\mathbf{x}}_i)-\varphi_i(\mathbf{r}_i)\le-\tilde\rho\|\boldsymbol{\xi}\|_2^2$ for $\boldsymbol{\xi}=\hat{\mathbf{x}}_i-\mathbf{r}_i$. With \eqref{eq:span_lemma} for $\mathbf{w}=\boldsymbol{\lambda}^\ast-\hat{\boldsymbol{\lambda}}$, $L_i^{\boldsymbol{\lambda}^\ast}(\hat{\mathbf{x}}_i)-L_i^{\boldsymbol{\lambda}^\ast}(\mathbf{r}_i)=\boldsymbol{\lambda}^{\ast\top}\boldsymbol{\xi}+\varphi_i(\hat{\mathbf{x}}_i)-\varphi_i(\mathbf{r}_i)\le\frac{\epsilon_i}{2}\|\boldsymbol{\xi}\|_1-\tilde\rho\|\boldsymbol{\xi}\|_2^2$. Using $\|\boldsymbol{\xi}\|_1\le2E_i/\Delta t$ gives the first term of \eqref{eq:vs_freeze}; using $\|\boldsymbol{\xi}\|_1\le\sqrt{|\Tset_i|}\|\boldsymbol{\xi}\|_2$, since $\boldsymbol{\xi}$ vanishes outside $\Tset_i$, and maximizing over $\|\boldsymbol{\xi}\|_2$ gives the second. (b) Let $\Tset_i=\Tset_a\cup\Tset_b$ be the split, with indicator vectors $\mathbf{1}_a$ and $\mathbf{1}_b$, and let $\theta=(\Delta t/E_i)\sum_{t\in\Tset_b}x(t)\in[0,1]$ for $\mathbf{x}\in\mathcal{X}_i$. Since EV $i$ can deliver all its energy in either set, $L_i^{S\mathbf{1}_b}(\mathbf{x})=\theta SE_i/\Delta t$ and $L_i^{S\mathbf{1}_a}(\mathbf{x})=(1-\theta)SE_i/\Delta t$. A rule that does not observe the prices returns the same schedule, or distribution of schedules, under both; hence, one loss is at least $SE_i/(2\Delta t)$; the forecast $\hat{\mathbf{w}}=\mathbf{0}$ with $\mathbf{w}\in\{\epsilon\mathbf{1}_a,\epsilon\mathbf{1}_b\}$ gives the second bound. If $\mathbf{x}$ minimizes $\hat{\mathbf{w}}^\top\mathbf{x}$ and $\mathbf{x}'$ minimizes $\mathbf{w}^\top\mathbf{x}$ over $\mathcal{X}_i$, then $L_i^{\mathbf{w}}(\mathbf{x})\le(\mathbf{w}-\hat{\mathbf{w}})^\top(\mathbf{x}-\mathbf{x}')\le E_i\,\mathrm{sp}_{\Tset_i}(\mathbf{w}-\hat{\mathbf{w}})/\Delta t$ by \eqref{eq:span_lemma}. (c) As in \eqref{eq:ev_x_proj_v_unscaled}, $\hat{\mathbf{x}}_i(\hat{\boldsymbol{\lambda}})=\Pi_{\mathcal{X}_i}\big((\tilde\rho\mathbf{r}_i-\hat{\boldsymbol{\lambda}}+\vartheta\mathbf{1}-\boldsymbol{\omega}_i+\beta\mathbf{x}_i^{\mathrm{p}})/(\tilde\rho+\beta)\big)$ for every $\vartheta$ by the span inequality, and the projection, which acts only on $\Tset_i$, is nonexpansive.\hfill$\blacksquare$

\noindent\emph{Proof of Corollary~\ref{cor:content}:} By Theorem~\ref{thm:byz_collusion}, every accepted broadcast is the unique certified one, and by Corollary~\ref{cor:detection}(c), every accepted upload is the prescribed update or recovery of its EV, a deterministic function of its declared constraints, its last accepted upload, and certified data, including the certified forecast $\hat{\boldsymbol{\lambda}}$ of \eqref{eq:ml_lambda_ev}; a frozen EV contributes its last accepted upload. Hence, if two strategies induce the same erasure pattern and declarations, induction over the iterations shows that the certified broadcasts, the accepted uploads, and the recovered schedules coincide. The right-hand sides of \eqref{eq:feasibility}--\eqref{eq:vs_freeze} depend only on certified data, declarations, and the erasure pattern; hence, the guarantees hold uniformly.\hfill$\blacksquare$

\bibliographystyle{IEEEtran}
\begingroup\renewcommand{\baselinestretch}{1.00}\bibliography{references}\endgroup

\begin{thebibliography}{10}
\providecommand{\url}[1]{#1}
\csname url@samestyle\endcsname
\providecommand{\newblock}{\relax}
\providecommand{\bibinfo}[2]{#2}
\providecommand{\BIBentrySTDinterwordspacing}{\spaceskip=0pt\relax}
\providecommand{\BIBentryALTinterwordstretchfactor}{4}
\providecommand{\BIBentryALTinterwordspacing}{\spaceskip=\fontdimen2\font plus
\BIBentryALTinterwordstretchfactor\fontdimen3\font minus
  \fontdimen4\font\relax}
\providecommand{\BIBforeignlanguage}[2]{{%
\expandafter\ifx\csname l@#1\endcsname\relax
\typeout{** WARNING: IEEEtran.bst: No hyphenation pattern has been}%
\typeout{** loaded for the language `#1'. Using the pattern for}%
\typeout{** the default language instead.}%
\else
\language=\csname l@#1\endcsname
\fi
#2}}
\providecommand{\BIBdecl}{\relax}
\BIBdecl

\bibitem{Needell2023BeneficialCharging}
Z.~Needell, W.~Wei, and J.~E. Trancik, ``Strategies for beneficial electric
  vehicle charging to reduce peak electricity demand and store solar energy,''
  \emph{Cell Rep. Phys. Sci.}, vol.~4, no.~3, p. 101287, 2023.

\bibitem{Zhan2025_LargeScaleEVUsage}
W.~Zhan, Y.~Liao, J.~Deng, Z.~Wang, and S.~Yeh, ``Large-scale empirical study
  of electric vehicle usage patterns and charging infrastructure needs,''
  \emph{npj Sustain. Mobil. Transp.}, vol.~2, p.~9, 2025.

\bibitem{Powell2022NatureEnergy}
S.~Powell, G.~V. Cezar, L.~Min, I.~L. Azevedo, and R.~Rajagopal, ``Charging
  infrastructure access and operation to reduce the grid impacts of deep
  electric vehicle adoption,'' \emph{Nat. Energy}, vol.~7, no.~10, pp.
  932--945, 2022.

\bibitem{HybridFleetLoadScheduling}
H.~Ayg{\"u}n, A.~S. Ero{\u{g}}lu, and C.~E. Y{\i}lmaz, ``An optimal hybrid
  management of electric vehicle fleet charging and load scheduling in active
  electric distribution system,'' \emph{IEEE Trans. Ind. Appl.}, vol.~60,
  no.~4, pp. 5304--5316, 2024.

\bibitem{TwoStageAggregatorReliability}
H.~Ayg{\"u}n, A.~S. Ero{\u{g}}lu, and C.~E. Y{\i}lmaz, ``A two-stage optimal
  electric vehicles charging methodology based on aggregators considering grid
  reliability and operational efficiency,'' \emph{IEEE Trans. Ind. Appl.},
  vol.~61, no.~2, pp. 940--954, 2025.

\bibitem{Kececi2021CheatingCustomers}
C.~Kececi, M.~Ismail, and E.~Serpedin, ``Analysis of {EV} charging coordination
  efficiency in presence of cheating customers,'' \emph{IEEE Access}, vol.~9,
  pp. 153\,666--153\,677, Nov. 2021.

\bibitem{Fahim2025DynamicPlanningGCNN}
S.~R. Fahim, R.~Atat, C.~Kececi, A.~Takiddin, M.~Ismail, K.~R. Davis, and
  E.~Serpedin, ``Dynamic spatio-temporal planning strategy of {EV} charging
  stations and {DG}s using {GCNN}-based predicted power demand,'' \emph{IEEE
  Trans. Intell. Transp. Syst.}, vol.~26, no.~4, pp. 4528--4542, Apr. 2025.

\bibitem{Alizadeh2025TII}
M.~Alizadeh, A.~M. Amani, L.~Meegahapola, M.~Jalili, and O.~Hill, ``Optimal
  integration of {EV} charging stations into distribution network planning and
  operation,'' \emph{IEEE Trans. Ind. Informat.}, vol.~21, no.~9, pp.
  7120--7130, Sep. 2025.

\bibitem{Alizadeh2026TTE}
M.~Alizadeh, A.~M. Amani, L.~Meegahapola, M.~Jalili, and O.~Hill, ``Optimal
  traffic-informed design and operation of {EV} charging stations in power
  distribution grids,'' \emph{IEEE Trans. Transport. Electrific.}, vol.~12,
  no.~3, pp. 4375--4387, Jun. 2026.

\bibitem{Alizadeh2025ETFG}
M.~Alizadeh, M.~H. Nikkhah, A.~M. Amani, L.~Meegahapola, and M.~Jalili,
  ``Optimal traffic-informed integration of {V2G}-enabled charging
  infrastructure in power distribution network,'' in \emph{Proc. IEEE Int.
  Conf. Energy Technol. Future Grids (ETFG)}, Wollongong, Australia, Dec. 2025.

\bibitem{ADMMFastChargeDegradation}
Y.~Zhou, T.~Xu, and D.~P. Mandic, ``{ADMM}-based coordination of electric
  vehicles in constrained distribution networks considering fast charging and
  degradation,'' \emph{IEEE Trans. Intell. Transp. Syst.}, vol.~22, no.~1, pp.
  565--578, 2021.

\bibitem{UncertainDepartureLoadFlattening}
C.~Wang, W.~Wei, Z.~Hu, and L.~Tang, ``Electric vehicle charging scheduling
  strategy for supporting load flattening under uncertain electric vehicle
  departures,'' \emph{J. Mod. Power Syst. Clean Energy}, vol.~11, no.~5, pp.
  1634--1645, 2023.

\bibitem{Nimalsiri2020SurveyDistributedCharging}
N.~I. Nimalsiri, C.~P. Mediwaththe, E.~L. Ratnam, M.~Shaw, D.~B. Smith, and
  S.~K. Halgamuge, ``A survey of algorithms for distributed charging control of
  electric vehicles in smart grid,'' \emph{IEEE Trans. Intell. Transp. Syst.},
  vol.~21, no.~11, pp. 4497--4515, Nov. 2020.

\bibitem{Boyd2011ADMM}
S.~Boyd, N.~Parikh, E.~Chu, B.~Peleato, and J.~Eckstein, ``Distributed
  optimization and statistical learning via the alternating direction method of
  multipliers,'' \emph{Found. Trends Mach. Learn.}, vol.~3, no.~1, pp. 1--122,
  2011.

\bibitem{Fard2023ISIE}
M.~Fakouri~Fard, X.~Huo, and M.~Liu, ``Exploration of for-purpose decentralized
  algorithmic cyber attacks in {EV} charging control,'' in \emph{Proc. IEEE
  Int. Symp. Ind. Electron. (ISIE)}, Helsinki-Espoo, Finland, Jun. 2023, pp.
  1--6.

\bibitem{FakouriFard2025Targeted}
M.~Fakouri~Fard and M.~Liu, ``Targeted algorithmic purpose-driven cyber attacks
  in distributed multi-agent optimization,'' arXiv:2511.13859, 2025.

\bibitem{PerezDiaz2020Catching}
A.~Perez-Diaz, E.~H. Gerding, and F.~McGroarty, ``Catching cheats: Detecting
  strategic manipulation in distributed optimisation of electric vehicle
  aggregators,'' \emph{J. Artif. Intell. Res.}, vol.~67, pp. 437--470, 2020.

\bibitem{Blanchard2017Krum}
P.~Blanchard, E.~M. {El Mhamdi}, R.~Guerraoui, and J.~Stainer, ``Machine
  learning with adversaries: {Byzantine} tolerant gradient descent,'' in
  \emph{Proc. Adv. Neural Inf. Process. Syst. (NeurIPS)}, 2017.

\bibitem{Yin2018TrimmedMean}
D.~Yin, Y.~Chen, K.~Ramchandran, and P.~Bartlett, ``{Byzantine}-robust
  distributed learning: Towards optimal statistical rates,'' in \emph{Proc.
  35th Int. Conf. Mach. Learn. (ICML)}, 2018, pp. 5650--5659.

\bibitem{Sundaram2019Adversarial}
S.~Sundaram and B.~Gharesifard, ``Distributed optimization under adversarial
  nodes,'' \emph{IEEE Trans. Autom. Control}, vol.~64, no.~3, pp. 1063--1076,
  Mar. 2019.

\bibitem{Yemini2025Resilient}
M.~Yemini, A.~Nedi{\'c}, A.~J. Goldsmith, and S.~Gil, ``Resilient distributed
  optimization for multiagent cyberphysical systems,'' \emph{IEEE Trans. Autom.
  Control}, vol.~70, no.~6, pp. 3952--3967, Jun. 2025.

\bibitem{Qu2026Resilient}
G.~Qu, L.~Zhang, H.~Jin, X.~Xu, J.~Xie, and Z.~Zhou, ``A resilient distributed
  charging scheduling strategy for electric vehicles under cyber-attacks,''
  \emph{Energies}, vol.~19, no.~18, p. Art. no. 4479, 2026.

\bibitem{Baza2019BlockchainCC}
M.~Baza, M.~Nabil, M.~Ismail, M.~Mahmoud, E.~Serpedin, and M.~A. Rahman,
  ``Blockchain-based charging coordination mechanism for smart grid energy
  storage units,'' in \emph{Proc. IEEE Int. Conf. Internet of Things
  (iThings)}, 2019, arXiv:1811.02001.

\bibitem{OCPP201}
{Open Charge Alliance}, ``Open charge point protocol ({OCPP}) 2.0.1,'' Open
  Charge Alliance, Arnhem, The Netherlands, 2020.

\bibitem{ISO15118_20}
{International Organization for Standardization}, ``{ISO 15118-20:2022---Road
  Vehicles---Vehicle to Grid Communication Interface---Part 20: 2nd Generation
  Network Layer and Application Layer Requirements},'' ISO, Geneva,
  Switzerland, 2022.

\bibitem{Acharya2020Cybersecurity}
S.~Acharya, Y.~Dvorkin, H.~Pand{\v{z}}i{\'c}, and R.~Karri, ``Cybersecurity of
  smart electric vehicle charging: A power grid perspective,'' \emph{IEEE
  Access}, vol.~8, pp. 214\,434--214\,453, 2020.

\bibitem{Johnson2022Review}
J.~Johnson, T.~Berg, B.~Anderson, and B.~Wright, ``Review of electric vehicle
  charger cybersecurity vulnerabilities, potential impacts, and defenses,''
  \emph{Energies}, vol.~15, no.~11, 2022, {Art. no.} 3931.

\bibitem{Lamport1982Byzantine}
L.~Lamport, R.~Shostak, and M.~Pease, ``The {Byzantine} generals problem,''
  \emph{ACM Trans. Program. Lang. Syst.}, vol.~4, no.~3, pp. 382--401, Jul.
  1982.

\bibitem{Bracha1987Reliable}
G.~Bracha, ``Asynchronous {Byzantine} agreement protocols,'' \emph{Inf.
  Comput.}, vol.~75, no.~2, pp. 130--143, Nov. 1987.

\bibitem{Abouyoussef2025SecureSemiDecentralizedDWC}
M.~Abouyoussef, M.~Ismail, and M.~F. Shaaban, ``Secure and semi-decentralized
  blockchain-based privacy-preserving networking strategy for dynamic wireless
  charging of {EV}s,'' \emph{IEEE Trans. Netw. Service Manag.}, vol.~22, no.~6,
  pp. 6083--6096, Dec. 2025.

\bibitem{Shafee2020LyingEV}
A.~A. Shafee, M.~M. Fouda, M.~M. E.~A. Mahmoud, A.~J. Aljohani, W.~Alasmary,
  and F.~Amsaad, ``Detection of lying electrical vehicles in charging
  coordination using deep learning,'' \emph{IEEE Access}, vol.~8, pp.
  179\,400--179\,414, 2020.

\bibitem{Fahim2024GAETTE}
S.~R. Fahim, R.~Atat, C.~Kececi, A.~Takiddin, M.~Ismail, K.~R. Davis, and
  E.~Serpedin, ``Graph autoencoder-based power attacks detection for resilient
  electrified transportation systems,'' \emph{IEEE Trans. Transport.
  Electrific.}, 2024.

\bibitem{Takiddin2021Poisoning}
A.~Takiddin, M.~Ismail, U.~Zafar, and E.~Serpedin, ``Robust electricity theft
  detection against data poisoning attacks in smart grids,'' \emph{IEEE Trans.
  Smart Grid}, vol.~12, no.~3, pp. 2675--2684, May 2021.

\bibitem{Duan2019Resilient}
J.~Duan and M.-Y. Chow, ``A resilient consensus-based distributed energy
  management algorithm against data integrity attacks,'' \emph{IEEE Trans.
  Smart Grid}, vol.~10, no.~5, pp. 4729--4740, Sep. 2019.

\bibitem{Cachin2001Broadcast}
C.~Cachin, K.~Kursawe, F.~Petzold, and V.~Shoup, ``Secure and efficient
  asynchronous broadcast protocols,'' in \emph{Advances in
  Cryptology---{CRYPTO} 2001}, ser. LNCS, vol. 2139.\hskip 1em plus 0.5em minus
  0.4em\relax Springer, 2001, pp. 524--541.

\bibitem{Gennaro2007DKG}
R.~Gennaro, S.~Jarecki, H.~Krawczyk, and T.~Rabin, ``Secure distributed key
  generation for discrete-log based cryptosystems,'' \emph{J. Cryptol.},
  vol.~20, no.~1, pp. 51--83, 2007.

\bibitem{Eckstein1992DR}
J.~Eckstein and D.~P. Bertsekas, ``On the {Douglas--Rachford} splitting method
  and the proximal point algorithm for maximal monotone operators,''
  \emph{Math. Program.}, vol.~55, no. 1--3, pp. 293--318, 1992.

\bibitem{BLS2004}
D.~Boneh, B.~Lynn, and H.~Shacham, ``Short signatures from the {Weil}
  pairing,'' \emph{Journal of Cryptology}, vol.~17, no.~4, pp. 297--319, 2004.

\bibitem{Boldyreva2003Threshold}
A.~Boldyreva, ``Threshold signatures, multisignatures and blind signatures
  based on the gap-{Diffie--Hellman}-group signature scheme,'' in \emph{Public
  Key Cryptography---{PKC} 2003}, ser. LNCS, vol. 2567.\hskip 1em plus 0.5em
  minus 0.4em\relax Springer, 2003, pp. 31--46.

\bibitem{Douceur2002Sybil}
J.~R. Douceur, ``The {Sybil} attack,'' in \emph{Peer-to-Peer Systems ({IPTPS}
  2002)}, ser. LNCS, vol. 2429.\hskip 1em plus 0.5em minus 0.4em\relax
  Springer, 2002, pp. 251--260.

\bibitem{pritchard2026evwatts}
\BIBentryALTinterwordspacing
E.~Pritchard, ``{EVWatts} public database,'' Apr. 2026, {U.S.} Department of
  Energy, Office of Scientific and Technical Information. Accessed: Apr. 15,
  2026. [Online]. Available: \url{https://www.osti.gov/biblio/1970735}
\BIBentrySTDinterwordspacing

\bibitem{EnergyOnlineCAISO}
\BIBentryALTinterwordspacing
{Energy Online}, ``{CAISO} historical system load and day-ahead market
  prices,'' Energy Online data portal, 2025, accessed: Apr. 15, 2026. [Online].
  Available: \url{https://www.energyonline.com/Data/Default.aspx}
\BIBentrySTDinterwordspacing

\bibitem{PGE_BEV_Tariff}
\BIBentryALTinterwordspacing
{Pacific Gas and Electric Company}, ``Business electric vehicle ({BEV}) rate
  plans,'' 2026, accessed: Apr. 15, 2026. [Online]. Available:
  \url{https://www.pge.com/en/account/rate-plans/business-ev-rate-plans.html}
\BIBentrySTDinterwordspacing

\bibitem{Ke2017LightGBM}
G.~Ke, Q.~Meng, T.~Finley, T.~Wang, W.~Chen, W.~Ma, Q.~Ye, and T.-Y. Liu,
  ``{LightGBM}: A highly efficient gradient boosting decision tree,'' in
  \emph{Proc. Adv. Neural Inf. Process. Syst. (NeurIPS)}, 2017, pp. 3146--3154.

\end{thebibliography}

\end{document}